\documentclass[11pt]{article} 
\usepackage[final]{acl}
\usepackage{hyperref}
\usepackage{url}
\usepackage{xurl}
\usepackage{booktabs}       
\usepackage{amsfonts}       
\usepackage{nicefrac}       
\usepackage{microtype}      

\usepackage{comment}
\usepackage{lipsum}
\usepackage{times}
\usepackage{latexsym}
\usepackage[T1]{fontenc}
\usepackage[utf8]{inputenc}
\usepackage{microtype}
\usepackage{inconsolata}
\usepackage{lipsum}
\usepackage{stfloats}
\usepackage{graphicx}
\usepackage{paralist}
\usepackage{amsmath}
\usepackage{multirow}
\usepackage{caption}
\usepackage{subcaption}
\usepackage{xspace}
\usepackage{makecell}
\usepackage{algorithm}  
\usepackage{algorithmic}
\usepackage{listings}
\usepackage{booktabs}
\usepackage{colortbl}
\usepackage{bbm}
\usepackage{amssymb}
\usepackage{float}
\usepackage[capitalise]{cleveref}
\usepackage{enumitem}
\usepackage{breakurl}
\usepackage{paralist}
\usepackage{fontawesome5}
\usepackage{academicons}

\definecolor{darkblue}{rgb}{0.0, 0.0, 0.5}

\hypersetup{
    colorlinks=true,
    linkcolor=darkblue,      
    citecolor=darkblue,      
    urlcolor=darkblue        
}

\title{Steering Knowledge Selection Behaviours in LLMs \\ via SAE-Based Representation Engineering}

\author{
\normalsize
Yu Zhao$^1$\quad
Alessio Devoto$^3$\quad
Giwon Hong$^1$\quad
Xiaotang Du$^1$\quad
Aryo Pradipta Gema$^1$ \\ \bf
\normalsize
Hongru Wang$^2$\quad
Xuanli He$^4$\quad
Kam-Fai Wong$^2$\quad
Pasquale Minervini$^{1,5}$\\ 
\normalsize
$^1$University of Edinburgh \quad
$^2$The Chinese University of Hong Kong \\ 
\normalsize
$^3$Sapienza University of Rome \quad
$^4$University College London \quad $^5$Miniml.AI\\
\normalsize
\texttt{\{yu.zhao, p.minervini\}@ed.ac.uk} \\
\footnotesize
{\faGithub}\hspace{0.5em}\texttt{\url{https://github.com/yuzhaouoe/SAE-based-representation-engineering}}
}

\DeclareMathOperator*{\argmin}{arg\,min}

\newcommand{\ConflictEvidence}{$E_C$\xspace}
\newcommand{\MemorisedEvidence}{$E_M$\xspace}

\newcommand{\ConflictAnswer}{$C$\xspace}
\newcommand{\MemorisedAnswer}{$M$\xspace}

\newcommand{\ConflictEvidenceDataset}{$\mathcal{D}_{E_C}$\xspace}
\newcommand{\MemorisedEvidenceDataset}{$\mathcal{D}_{E_M}$\xspace}

\newcommand{\UseContextDataset}{$\mathcal{D}_C$\xspace}
\newcommand{\UseParameterDataset}{$\mathcal{D}_M$\xspace}

\newcommand{\MethodName}{\textsc{SpARE}\xspace}

\usepackage{xspace}
\newcommand*{\eg}{e.g.\@\xspace}

\newcommand{\emc}{$\text{EM}_{C}$\xspace}
\newcommand{\emm}{$\text{EM}_{M}$\xspace}
\newcommand{\emctom}{$\text{EM}_{C \rightarrow M}$\xspace}
\newcommand{\emmtoc}{$\text{EM}_{M \rightarrow C}$\xspace}
\newcommand{\emmtom}{$\text{EM}_{M \rightarrow M}$\xspace}
\newcommand{\emctoc}{$\text{EM}_{C \rightarrow C}$\xspace}

\begin{document}

\maketitle

\begin{abstract}
Large language models (LLMs) can store a significant amount of factual knowledge in their parameters. However, their parametric knowledge may conflict with the information provided in the context---this phenomenon, known as \emph{context-memory knowledge conflicts}~\footnote{We will refer to these as \emph{knowledge conflicts} for brevity.}, can lead to undesirable model behaviour, such as reliance on outdated or incorrect information. Analysing the internal activations of LLMs, we find that they can internally register the signals of knowledge conflict at mid-layers. Such signals allow us to detect whether a knowledge conflict occurs and use \emph{inference-time} intervention strategies to resolve it. In this work, we propose \MethodName, a \emph{training-free} representation engineering method that uses pre-trained sparse auto-encoders (SAEs) to control the knowledge selection behaviour of LLMs. \MethodName identifies the functional features that control the knowledge selection behaviours and applies them to edit the internal activations of LLMs at inference time. Our experimental results show that \MethodName can effectively control the usage of either knowledge source to resolve knowledge conflict in open-domain question-answering tasks, surpassing existing representation engineering methods ($+10\%$) as well as contrastive decoding methods ($+15\%$).
\end{abstract}

\section{Introduction}

\begin{figure}[t]
    \centering
    \includegraphics[width=\columnwidth]{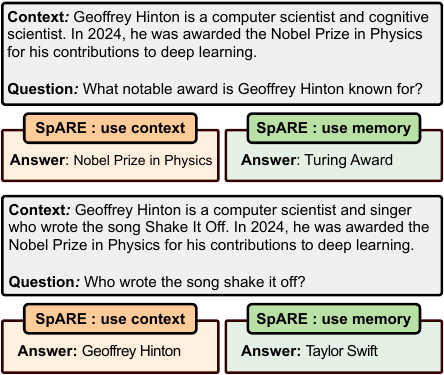}
    \caption{
    In the event of a knowledge conflict, the model can rely on the context or on the parametric knowledge.
    The figure presents the predictions of Llama2-7B steered by \MethodName.
    }
    \vspace{-10pt}
    \label{fig:first_page}
\end{figure}
Large language models (LLMs) have shown remarkable capability to memorise factual knowledge and solve knowledge-intensive tasks~\citep{DBLP:conf/emnlp/PetroniRRLBWM19,gpt3,llama2,mistral,gemini}.
Nevertheless, the knowledge stored in their parameters (\emph{parametric knowledge}) can be inaccurate or outdated~\citep{xu2024knowledge}.
To alleviate this issue, retrieval and tool-augmented approaches have been widely adopted to provide LLMs with external knowledge (\emph{contextual knowledge})~\citep{DBLP:conf/emnlp/KarpukhinOMLWEC20,lewis2020retrieval,DBLP:conf/emnlp/WuZHMS022,schick2024toolformer}. 
However, contextual knowledge may sometimes conflict with the parametric knowledge of the model, leading to what we refer to as \emph{knowledge conflicts}.
Such conflicts can cause undesired behaviour, where the model may rely on inaccurate information sources, resulting in incorrect outputs~\citep{when-not-to-trust-language-models,conflictqa,conflictbank,resolving-knowledge-conflict, zhao2024attacks}.
Prior research found that LLMs tend to prefer contextual knowledge (\eg, retrieved passages) over their parametric knowledge when conflicts occur~\citep{conflictbank,conflictqa,macnoise}.
For instance, \citet{conflictbank} show that most LLMs choose parametric knowledge in less than $10\%$ examples.
However, in more general applications, LLMs should retain the ability to use their parametric knowledge when presented with misinformation~\citep{chen2023combating,chen2023can,zou2024poisonedrag,when-not-to-trust-language-models,zhong2023poisoning}.
Existing works investigate fine-tuning and prompting-based strategies to detect and resolve knowledge conflicts~\citep{resolving-knowledge-conflict}; however, they need additional interactions with the model, \eg, by asking the LLMs to examine the conflicts sentence by sentence, resulting in high latency times and preventing practical applications.

In this work, we investigate \emph{representation engineering} methods to efficiently steer the usage of parametric and contextual knowledge of LLMs at \emph{inference time}.
Although representation engineering has provided an efficient and transparent framework for controlling the behaviour of LLMs, we find that existing methods fail to effectively steer knowledge usage.
This may be because these methods directly modify the internal activations of LLMs, such as hidden states~\citep{actadd, representation-engineering} or MLP activations~\citep{spectral-editing,rome}. These activations are polysemantic dense vectors that overlap with many independent semantic features~\citep{olah_distributed_2023}. 
Thus, minor edits in one dimension can influence multiple semantic features, making it difficult to adjust activations accurately without affecting other features in practice.

Recently, sparse auto-encoders (SAEs) have been proposed to address the difficulty of interpreting polysemantic activations by decomposing them into a large-scale monosemantic feature dictionary~\citep{cunningham2024sparse,scaling-and-evaluating-sae,scaling-monosemanticity}.
Therefore, we introduce SAEs as a tool for precise activation editing to guide the knowledge selection of LLMs.
Specifically, we propose \MethodName, a \textbf{Sp}arse \textbf{A}uto-Encoder-based \textbf{R}epresentation \textbf{E}ngineering method to steer the knowledge selection behavior of LLMs.
\MethodName first identifies the SAE activations that are related to specific knowledge selection behaviours (\cref{sec:measure-correlation}); then, it extracts functional features that control the usage of contextual and parametric knowledge, and finally applies them to steer the behaviour of the model (\cref{sec:activation-editing}).
Our experimental results on open-domain question-answering tasks show that \MethodName effectively controls the knowledge selection behaviours by utilising a small set of SAE features, e.g., less than $0.05\%$ SAE activations for Gemma2-9B in the $6$ layers\footnote{For Gemma2-9B, we use the pre-trained SAEs from GemmaScope \url{https://huggingface.co/google/gemma-scope}, and the selected activations is presented in~\cref{sec:gemma-selected-activations}.}.
\MethodName yields more accurate results than state-of-the-art representation engineering methods ($+10\%$), contrastive decoding ($+15\%$), and in-context learning ($+7\%$), achieving the best performance on steering knowledge selection behaviours of LLMs under knowledge conflicts.

\section{Background}

\paragraph{Problem Setup}
Following \citet{nqswap,macnoise,conflictqa}, we use open-domain question-answering (ODQA) tasks to investigate the behaviour of LLMs when there is a conflict between the parametric knowledge of the model and contextual knowledge.
In ODQA datasets with knowledge conflicts, each instance is presented as $(Q, E_M, M, E_C, C)$, where $Q$ is the question, \MemorisedEvidence is the evidence that supports the memorised knowledge stored in the model parameters, \ConflictEvidence is the evidence that conflicts with the language model's memorised knowledge, \MemorisedAnswer is the answer based on the \MemorisedEvidence, and \ConflictAnswer is the answer based on the \ConflictEvidence.

\paragraph{Sparse Auto-Encoders}
Recent works have proposed using sparse auto-encoders (SAEs) to interpret the complex representations of LLMs by decomposing them into a large set of monosemantic features~\cite{cunningham2024sparse,scaling-and-evaluating-sae,scaling-monosemanticity}.
Given an activation $\mathbf{h}\in \mathbb{R}^{d}$ from the residual stream of LLMs, a SAE with $n$ latent dimensions encodes it into a sparse vector $\mathbf{z}\in \mathbb{R}^n$ and decodes it to recover $\mathbf{h}$:
\begin{equation} \label{eq:sae}
\begin{aligned}
%
& 
f_{\theta}(\mathbf{h}) = \sigma\left( \mathbf{W}_{\theta} \left(\mathbf{h} - \mathbf{b} \right) + \mathbf{b}_{\theta} \right) = \mathbf{z},\\
& 
g_{\phi}(\mathbf{z}) = \mathbf{W}_{\phi}\ 
 \mathbf{z} =\sum_{i=1}^{n}z_i \mathbf{f}_i + \mathbf{b} = \hat{\mathbf{h}}
\end{aligned}
\end{equation}
\noindent where $\sigma$ is an activation function that outputs a non-negative value such as ReLU, $\mathbf{W}_{\theta} \in \mathbb{R}^{n\times d}$, $\mathbf{b} \in \mathbb{R}^d$, $\mathbf{b}_{\theta} \in \mathbb{R}^n$, $\mathbf{W}_{\phi} \in \mathbb{R}^{d\times n}$, $z_i$ is the $i$-th element of the SAE activation $\mathbf{z}$, and $\mathbf{f}_i\in \mathbb{R}^d$ is the $i$-th column of $\mathbf{W}_{\phi}$.
The $\{\mathbf{f}_i\}_{i=1}^{n}$ learned through the SAE are considered highly monosemantic, and the SAE activation $\mathbf{z}$ indicates the activated values of $\{\mathbf{f}_i\}_{i=1}^{n}$.

\section{Detection of Knowledge Conflicts} \label{sec:detection}

\label{sec:knowledge-conflict-detection}
In this section, we investigate whether we can detect the occurrence of conflicts during the generation process, since identifying such conflicts is a prerequisite for exploring inference-time strategies to control the LLM.
We focus on the residual stream~\citep{mathematical-framework} of the model and look for a signal of knowledge conflict.
To this end, we create two groups of input instances, \MemorisedEvidenceDataset$=\{(Q, E_M)\}$ and \ConflictEvidenceDataset$=\{(Q, E_C)\}$.
In \MemorisedEvidenceDataset, the model is provided with a context that is coherent with the model internal memorized knowledge, whereas in \ConflictEvidenceDataset the model is provided with a context that does not agree with model parametric knowledge, thus causing a knowledge conflict.
To determine whether a signal of conflict arises in the residual stream, we focus on the last position of the sequence during generation, which is supposed to encode the information to predict the first token of the answer.

\begin{figure}[t]
        \centering
    \begin{subfigure}[b]{0.49\linewidth}
        \centering
        \includegraphics[width=\linewidth]{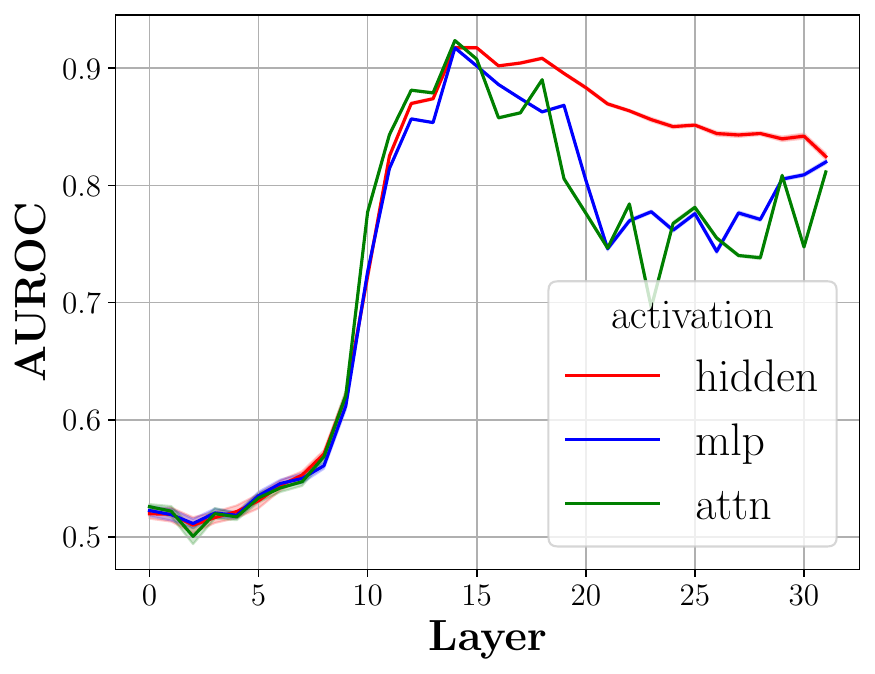}
        \caption{Llama2-7B}
    \end{subfigure}
    \begin{subfigure}[b]{0.49\linewidth}
        \centering
        \includegraphics[width=\linewidth]{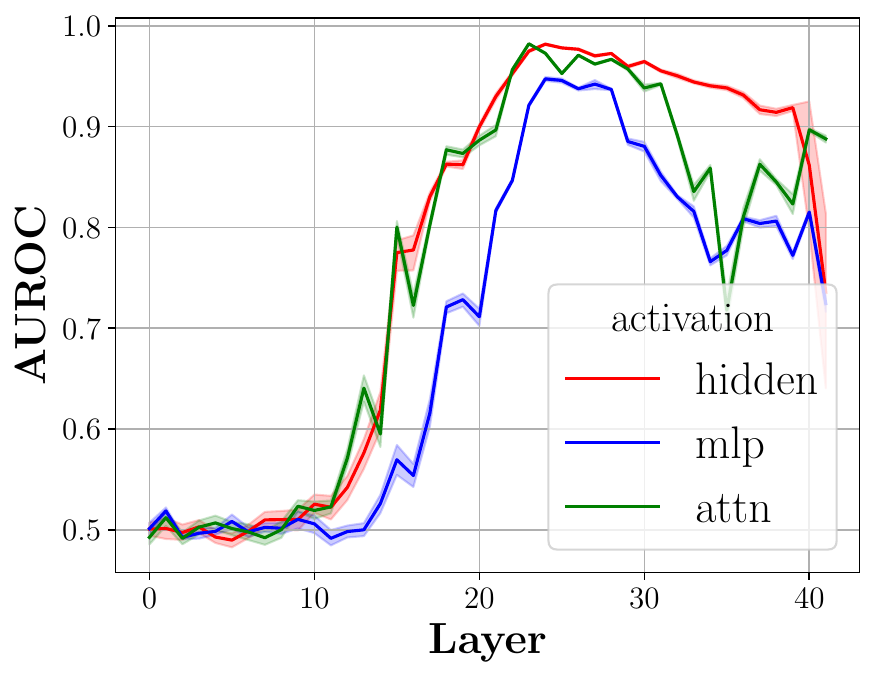}
        \caption{Gemma2-9B}
    \end{subfigure}
\caption{The knowledge conflict probing results of Llama2-7B and Gemma2-9B on NQSwap \citep{nqswap}.
The probing results on hidden states, MLP and Self-Attention activations are coloured differently.
}
\label{fig:knowledge-conflict-probing}
\end{figure}

We apply a linear probing method~\citep{conneau2018you,zhu2023physics,allen2023physics} to investigate whether the residual stream contains a signal of knowledge conflict.
Specifically, we train logistic regression models to classify whether a given activation (the hidden state, MLP or Self-Attention activations) is from the \ConflictEvidenceDataset or \MemorisedEvidenceDataset, i.e. whether it contains a knowledge conflict or not.
We use activations from each layer as input and formulate this as a binary classification task. The evaluation is conducted on a held-out test set.
We present probing results on Llama2-7B~\citep{llama2} and Gemma2-9B~\citep{gemma2} using AUROC as metric in~\cref{fig:knowledge-conflict-probing}.
We observe that the probing accuracy increases from the first layer to the middle layers, and this trend is the same across different types of activations.
This indicates that we can detect the signal of knowledge conflict in the residual stream of the mid-layers.
The probing accuracy decreases in the later layers, especially for MLP and Self-Attention activations, which indicates that MLP and Self-Attention modules do not further add the signal of conflicting knowledge to the residual stream.
We provide more details and analysis about knowledge conflict detection in~\cref{sec:more-knowledge-conflict-detection} and~\citet{residual-stream-knowledge-conflict}.

The above analysis shows that knowledge conflicts can be identified in the internal states of LLMs. 
Moreover, it provides insight into which layers can be more influential in the knowledge selection (\cref{sec:edit-layers}).

\section{Resolving Knowledge Conflicts by Representation Engineering} \label{sec:spare}
In this section, we introduce \MethodName, our SAE-based representation engineering method, to steer the usage of parametric and contextual knowledge to generate the answers.
\MethodName consists of the three following steps: \textit{1)}~collecting activations that lead to different knowledge selection behaviours (\cref{sec:prepare-activation}); \textit{2)}~identifying SAE activations that are related to each knowledge selection behaviour (\cref{sec:measure-correlation}); \textit{3)}~steering the usage of either knowledge source by editing the hidden states of LLMs at inference time (\cref{sec:activation-editing}).

\subsection{Collecting Activations with Different Knowledge Selection Behaviours}
\label{sec:prepare-activation}
We showed in \cref{sec:knowledge-conflict-detection} that we can detect the knowledge conflict by probing the residual stream. We now want to characterise the activations that lead to different knowledge selection behaviours.  
To this end, given a set of instances \ConflictEvidenceDataset that cause a knowledge conflict, we separate it into two groups based on the model's predictions: \UseContextDataset, where the model generates an answer that aligns with the context, and \UseParameterDataset, where the model ignores the context and generates an answer relying on the parametric knowledge.
These two subsets characterise two knowledge selection behaviours of the model.
In the following, we omit the notation to specify the layer of $\mathbf{h}$ and $\mathbf{z}$ for simplicity, as the method can be applied to arbitrary layers.
We collect the hidden state at the last position of the input that is used to generate the first token of the answer.

We collect the hidden states from \UseContextDataset and \UseParameterDataset for $N$ samples, denoting them as $\{\mathbf{h}^j_{C}\}_{j=1}^N$ and $\{\mathbf{h}^j_{M}\}_{j=1}^N$, respectively. 
We then obtain the SAE activation for each sample by $\mathbf{z}^j_{C} = f_{\theta}(\mathbf{h}^j_C)$ and $\mathbf{z}^j_{M} = f_{\theta}(\mathbf{h}^j_M)$.
Finally, we compute the average of the sets $\{\mathbf{z}_{C}^j\}_{j=1}^N$ and $\{\mathbf{z}_{M}^j\}_{j=1}^N$ to obtain the mean vectors $\overline{\mathbf{z}_C}$ and $\overline{\mathbf{z}_M}$, respectively.
More details are presented in~\cref{sec:collecting-activation-details} and~\cref{sec:hidden-num}.
At this stage, $\overline{\mathbf{z}_C}$ and $\overline{\mathbf{z}_M}$ contain the information to steer the generation towards $C$ or $M$.
However, there might still be instance-specific activations with non-zero values that are not responsible for the knowledge selection behaviour.
In the next section, we identify functional activations related to knowledge selection behaviours and then construct two orthogonal SAE activations, ${\mathbf{z}_C}$ and $\mathbf{z}_M$ for steering the knowledge selection behaviours.

\begin{figure}[t]
\centering
\includegraphics[width=0.85\linewidth]{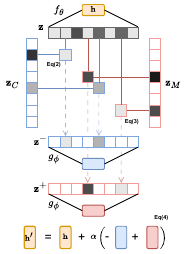}
\caption{
The workflow of \MethodName steers the knowledge selection behaviour.
The figure presents an example of steering the model to use parametric knowledge.
First, the SAE encoder $f_\theta$ encodes hidden state $\mathbf{h}$ into the SAE activation $\mathbf{z}$.
Then, it determines the values of SAE activations $\mathbf{z}^{-}$ and $\mathbf{z}^{+}$ for editing (\cref{eq:remove-act} and \cref{eq:add-act}).
Finally, we edit the hidden state using the features extracted from the SAE decoder $g_\phi$ (\cref{eq:hidden-edit}).
}
\label{fig:method}
\end{figure}

\subsection{Identifying Functional SAE Activations}
\label{sec:measure-correlation}
As shown by previous works~\citep{scaling-and-evaluating-sae,scaling-monosemanticity}, a single SAE activation can capture one monosemantic feature.
In this work, we hypothesise a combination of a small set of SAE activations can be responsible for a functional feature, such as knowledge selection in case of conflict.
Our hypothesis is motivated by Task Vector~\citep{task-vector,DBLP:conf/iclr/ToddLSMWB24}, which shows that hidden states contain the functional information that drives a task. 

We now show how we find the SAE activations that are responsible for driving the knowledge selection. 
First, we calculate mutual information between each SAE activation and the knowledge selection behaviours, which measures to which extent the behaviour depends on each activation.
Let the random variable $Z_i$ be the $i$th activation of SAE, and $Y=\{C, M\}$ be the generated answers; we calculate the mutual information $I(Z_i; Y)$ between them.
A higher $I(Z_i;Y)$ indicates a higher dependency between $Z_i$ and the knowledge selection behaviour.
We then select the top-$k$ activations with the highest $I(Z_i;Y)$, denoted as $\mathcal{Z}$.
More details are available in~\cref{sec:detail-indentify-functional-activations}

In the following, we determine which knowledge selection behaviour each $Z_i \in \mathcal{Z}$ positively correlates with.
Given the sets of activations $\{\mathbf{z}_{C}^j\}_{j=1}^N$ and $\{\mathbf{z}_{M}^j\}_{j=1}^N$, we estimate the expected value of each activation feature $Z_i \in \mathcal{Z}$ in both sets, denoted as $\mathbb{E}_{C}[Z_i]$ and $\mathbb{E}_{M}[Z_i]$.
We then have that $Z_i$ is positively correlated with the behaviour of selecting contextual knowledge if $\mathbb{E}_{C}[Z_i] - \mathbb{E}_{M}[Z_i] > 0$. Conversely, if this condition is not met, $Z_i$ is positively correlated with the behaviour of selecting parametric knowledge.
Finally, we construct two functional SAE activations $\mathbf{z}_C$ and $\mathbf{z}_M \in \mathbb{R}^n$, that steer the usage of contextual and parametric knowledge, respectively.
For each element, $z_{Ci}$ and $z_{Mi}$ are set to $0$ if $Z_i \notin \mathcal{Z}$, and the remaining values are taken from $\overline{\mathbf{z}_{C}}$ and $\overline{\mathbf{z}_{M}}$ based on their expectations:
\begin{align*}
z_{Ci} = &
\begin{cases}
\overline{z_{C}}_i, & \text{ if } \mathbb{E}_{C}[Z_i] - \mathbb{E}_{M}[Z_i] > 0 \\
0, & \text{otherwise}
\end{cases} \\
z_{Mi} = &
\begin{cases}
\overline{z_{M}}_i, & \text{ if } \mathbb{E}_{C}[Z_i] - \mathbb{E}_{M}[Z_i] < 0 \\
0, & \text{otherwise}
\end{cases}
\end{align*}

\subsection{Editing Activations to Steer Behaviours}
\label{sec:activation-editing}
In the following, we introduce how we utilise the functional activation $\mathbf{z}_C$ and $\mathbf{z}_M$ to control the usage of knowledge sources at inference time.
Suppose we want to control the LLM to use its parametric knowledge and ignore the conflict contextual knowledge that might be misinformation.
In this case, we aim to \emph{remove} the features that steer the contextual knowledge usage and \emph{add} the features that steer the parametric knowledge usage.
To avoid removing or adding unnecessary features, we restrict the values to edit by the following two constraints.
Let an activation be $\mathbf{h}$ and corresponding SAE activations  be $\mathbf{z}=f_{\theta}(\mathbf{h})$.
First, in \cref{eq:remove-act}, we determine the value we need to remove from $z_i$ to avoid the undesired behaviour, i.e., generating contextual knowledge in this case.
At this step, we ensure that the resulting activation remains non-negative after the removal, i.e., subtract at most $z_i$ when $z_i < z_{Ci}$:
\begin{equation}
\label{eq:remove-act}
z_{i}^{-} = \min\left\{z_{i}, z_{Ci}\right\}.
\end{equation}
Then, in~\cref{eq:add-act}, we determine the value we need to add to $z_i$ to encourage the desired behaviour, i.e., generating parametric knowledge in this case.
Here, we ensure that no excess value is added once the activation reaches $z_{Mi}$:
\begin{equation}
\label{eq:add-act}
z_i^{+} = \max\left\{z_{Mi} - z_{i}, 0\right\}.
\end{equation}
Finally, we obtain the edited hidden states $\mathbf{h}'$ by:
\begin{equation}
\label{eq:hidden-edit}
\mathbf{h}' = \mathbf{h} + \alpha \left(
- g_\phi\left(\mathbf{z}^{-}\right)
+ g_\phi\left(\mathbf{z}^{+}\right)
\right),
\end{equation}
\noindent where $\alpha \in \mathbb{R}^+$ is a user-defined hyperparameter that controls the degree of editing.
Note that we do not directly edit the activation $\mathbf{z}$ of hidden state $\mathbf{h}$ to obtain the modified hidden states by
$\mathbf{h}'=g_{\phi}(\mathbf{z}')$
with $\mathbf{z}' = \mathbf{z} - \mathbf{z}^{-} + \mathbf{z}^{+}$
for two reasons: \textit{1)}~it can result in unexpected information loss of original $\mathbf{h}$ due to the reconstruction loss of the SAE, leading to a nearly zero accuracy in our experiments; \textit{2)}~the definition of the SAE activation shown in~\cref{eq:sae} requires each element of $\mathbf{z}$ be a non-negative value, which prevents us from using $\alpha$ to flexibly control the strengthen of editing like~\cref{eq:hidden-edit}.

Similarly, if we want to control the model to be faithful to the context in the case of the contextual knowledge is more likely correct, we can swap $z_{Ci}$ and $z_{Mi}$ in \cref{eq:remove-act} and \cref{eq:add-act}.
In this work, we only edit hidden states at the last position of the input for ODQA tasks.

\section{Experimental Results}
\label{sec:experiments}

\subsection{Settings}

\paragraph{Datasets}
We use two widely adopted open-domain question-answering datasets with knowledge conflicts NQSwap~\citep{nqswap} and Macnoise~\citep{macnoise} to investigate the controlling capability of several methods.

\paragraph{Models}
We evaluate our method using Llama3-8B~\citep{llama3} and Gemma2-9B~\citep{gemma2}, which have corresponding public pre-trained SAEs.
Moreover, we also evaluate our method using Llama2-7B~\citep{llama2} with our pre-trained SAEs to examine the feasibility of adopting \MethodName to an LLM without public SAEs.
More details are presented in~\cref{sec:details-of-sae}.
\paragraph{Evaluation}
We use the greedy decoding method to evaluate the LLMs in an open-ended generation setting.
We use $3$ in-context demonstrations to align the answer format. 
The demonstrations use non-conflict evidence \MemorisedEvidence and memorised answer \MemorisedAnswer, so they do not point out which knowledge source to select.
More details are presented in~\cref{sec:development-set-and-demonstrations}.
The test examples use \ConflictEvidence, leading to a knowledge conflict for LLMs, and a behaviour-controlling method needs to steer the usage of either parametric or contextual knowledge to generate the answer.
We compare the evaluation results under control with the results without any control to show each method's controlling capability.

\begin{table*}[t]
\centering
\small
\resizebox{\linewidth}{!}{
\begin{tabular}{llcccccc} 
\toprule
\multirow[m]{2}{*}{\bf Metric} & \multirow[m]{2}{*}{\bf Method} & \multicolumn{3}{c}{\textbf{NQSwap}~\citep{nqswap}} & \multicolumn{3}{c}{\textbf{ Macnoise}~\citep{macnoise}} \\ \cmidrule(lr){3-5} \cmidrule(lr){6-8}
& & \bf Llama3-8B & \bf Llama2-7B & \bf Gemma-2-9B & \bf Llama3-8B & \bf Llama2-7B & \bf Gemma-2-9B \\
\midrule

\multicolumn{2}{l}{\textbf{Steer to Use Parametric Knowledge}} \\\cmidrule{1-2}
\rowcolor{gray!30} %
\multirow[m]{9}{*}{\emm}
& \emph{Without Controlling} & $26.63_{\pm 6.02}$ & $22.23_{\pm 4.75}$ & $26.32_{\pm 1.80}$ & $18.96_{\pm 2.65}$ & $22.37_{\pm 1.89}$ & $17.06_{\pm 3.79}$  \\
& TaskVec~\citep{task-vector} & $24.16_{\pm 6.58}$ & $24.88_{\pm 0.85}$ & $29.85_{\pm 0.83}$  & $21.23_{\pm 1.89}$  & $22.93_{\pm 2.31}$ &  $28.92_{\pm 1.19}$  \\
& ActAdd~\citep{actadd} &  $37.87_{\pm 8.96}$  & $31.43_{\pm 3.68}$ & $27.67_{\pm 0.82}$ &  $26.17_{\pm0.22}$ & $27.52_{\pm 3.07}$ & $29.75_{\pm 1.68}$  \\
& SEA$_{\text{linear}}$~\citep{spectral-editing} & $21.03_{\pm 1.83}$ & $23.73_{\pm 0.86}$  & $24.43_{\pm 0.91}$ &  $12.84_{\pm 0.18}$ & $15.64_{\pm 0.24}$ & $28.10_{\pm 2.78}$  \\

& SEA$_{\text{SqExp}}$~\citep{spectral-editing} & $13.64_{\pm 1.62}$ & $16.66_{\pm 0.55}$  & $23.79_{\pm 1.38}$ &  $14.24_{\pm 1.45}$ & $16.24_{\pm 1.06}$ & $28.07_{\pm 1.30}$  \\

& DoLa~\citep{dola} & $25.53_{\pm 5.19}$ & $16.50_{\pm 3.91}$ & $20.58_{\pm 1.06}$ & $16.52_{\pm 2.65}$ & $15.66_{\pm 0.88}$ & $19.81_{\pm 2.58}$ \\
& $^\flat$CAD~\citep{cad} & $33.72_{\pm 0.84}$ & $31.23_{\pm 1.45}$ & $41.17_{\pm 0.59}$ & $28.58_{\pm 0.75}$ & $30.81_{\pm 0.94}$ & $\underline{33.15}_{\pm 2.87}$ \\
& $^\sharp$ICL~\citep{gpt3} & $\underline{43.73}_{\pm 1.55}$ & $\underline{31.67}_{\pm 5.49}$ & $\underline{43.10}_{\pm 3.63}$ & $\underline{29.54}_{\pm 4.16}$ & $\underline{31.23}_{\pm 0.94}$ & $21.91_{\pm 2.35}$ \\
& \MethodName (Ours) & $\textbf{47.51}_{\pm 1.30}$ & $\textbf{43.76}_{\pm 3.14}$ & $\textbf{44.11}_{\pm 1.30}$ & $\textbf{30.72}_{\pm 1.42}$ & $\textbf{35.43}_{\pm 1.10}$ & $\textbf{35.53}_{\pm 2.07}$
\\ \midrule
\multicolumn{2}{l}{\textbf{Steer to Use Contextual Knowledge}} \\ \cmidrule{1-2}
\rowcolor{gray!30}
\multirow[m]{9}{*}{\emc} 
& \emph{Without Controlling} & $42.69_{\pm 8.40}$ & $41.67_{\pm 4.66}$ & $45.96_{\pm 2.48}$ & $69.36_{\pm 3.57}$ & $62.38_{\pm 3.05}$ & $59.25_{\pm 2.82}$ \\
& TaskVec~\citep{task-vector} & $41.88_{\pm 9.45}$ & $38.25_{\pm 1.23}$ & $45.52_{\pm 1.06}$ & $\underline{88.47}_{\pm 1.93}$  & $\underline{86.91}_{\pm 0.44}$ & $59.25_{\pm 1.49}$   \\
& ActAdd~\citep{actadd} & $51.91_{\pm 8.03}$ & $47.48_{\pm 3.93}$  & $46.90_{\pm 1.89}$ &  $73.01_{\pm 1.58}$ & $69.64_{\pm 0.20}$ & $59.66_{\pm 2.89}$  \\
& SEA$_{\text{linear}}$~\citep{spectral-editing} & $43.61_{\pm 10.3}$ & $47.73_{\pm 0.43}$  & $52.95_{\pm 1.90}$ &  $69.78_{\pm 0.97}$ & $67.32_{\pm 0.28}$ & $60.31_{\pm 2.25}$  \\

& SEA$_{\text{SqExp}}$~\citep{spectral-editing}  & $57.08_{\pm 2.92}$ & $48.04_{\pm 0.45}$  & $61.45_{\pm 0.54}$ &  $72.04_{\pm 1.60}$ & $68.20_{\pm 1.10}$ & $61.45_{\pm 0.30}$  \\

& DoLa~\citep{dola} & $44.29_{\pm 8.46}$ & $33.54_{\pm 3.38}$ & $15.90_{\pm 10.1}$ & $68.45_{\pm 3.83}$ & $50.95_{\pm 5.15}$ & $23.34_{\pm 10.5}$ \\
& $^\flat$CAD~\citep{cad} & $65.65_{\pm 5.50}$ & $54.69_{\pm 3.25}$ & $63.10_{\pm 2.32}$ & $78.69_{\pm 3.85}$ & $70.07_{\pm 3.77}$ & $\underline{64.12}_{\pm 4.44}$ \\
& $^\sharp$ICL~\citep{gpt3} & $\underline{73.35}_{\pm 3.82}$ & $\underline{63.33}_{\pm 3.50}$ & $\underline{70.19}_{\pm 2.51}$ & $51.75_{\pm 5.60}$ & $47.51_{\pm 1.86}$ & $47.24_{\pm 3.81}$ \\
& \MethodName (Ours) & $\textbf{77.69}_{\pm 1.24}$ & $\textbf{69.32}_{\pm 1.26}$ & $\textbf{73.78}_{\pm 0.74}$ & $\textbf{92.24}_{\pm 0.49}$ & $\textbf{87.30}_{\pm 1.96}$ & $\textbf{87.96}_{\pm 1.85}$
\\ \bottomrule
\end{tabular}
}
\caption{
Overall performance of steering the utilisation of parametric and contextual knowledge, measured by \emm and \emc.
"Without Controlling" indicates the baseline that we do not use any controlling methods to steer the generation.
$^\sharp$ICL is not an inference-time controlling strategy, which controls the behaviours by changing demonstrations. 
$^\flat$CAD needs additional forwarding for contrastive decoding.
}
\label{tab:main-results}
\end{table*}

\paragraph{Baselines}
We compare \MethodName against the following inference-time \emph{representation engineering} methods:
\begin{inparaenum}[1)]
\item TaskVec~\citep{task-vector};
\item ActAdd~\citep{actadd};
\item SEA~\citep{spectral-editing} with linear and non-linear versions, noted by subscript "linear" and "SqExp".
\end{inparaenum}
We compare with the following \emph{contrastive decoding} methods:
\begin{inparaenum}[1)]
\item DoLa~\citep{dola};
\item CAD~\citep{cad}.
\end{inparaenum}
Moreover, we also compare using in-context learning (ICL)~\citep{gpt3} to steer the knowledge selection. We use \ConflictEvidence and \ConflictAnswer in the demonstrations to guide the model to ignore its parametric knowledge and use the contextual knowledge, and use \ConflictEvidence and \MemorisedAnswer to guide the model to ignore the contextual knowledge and use its parametric knowledge.
ICL is not an inference-time strategy because it requires changing the original input of the model to achieve a desired behaviour.
More details of baseline implementation and hyperparameters searching are presented in~\cref{sec:re-baseline-details} and ~\cref{sec:baseline-details}.

\paragraph{Hyperparameters}
We select the proper hyperparameters for \MethodName in the developments set that is also used to select the hyperparameters of baselines, and the details are presented in~\cref{sec:baseline-details}.
In the following, we apply \MethodName from the 12th to the 15th and 13th to the 16th layer for Llama2-7B and Llama3-8B and from the 23rd to 25th and the 29th to 31st layers for Gemma2-9B; we analyse the performance of editing individual layers in \cref{sec:edit-layers}.

\subsection{Overall Performance Comparison}
\label{sec:overall-performance}
\paragraph{Metrics}
We use Exact Match (EM) to evaluate the performance.
Specifically, we evaluate the control capability of generating contextual or parametric answers using the following metrics: 
\begin{description}[leftmargin=*,nosep,topsep=0pt]
\item[\emc] accuracy of steering the usage of contextual knowledge to generate answers \ConflictAnswer.
\item[\emm] accuracy of steering the usage of parametric knowledge to generate answer \MemorisedAnswer.
\end{description}
\paragraph{Experimental Results} We present the main results in~\cref{tab:main-results}.
First, we find \emph{\MethodName outperforms existing representation engineering methods} TaskVec, ActAdd and SEA on steering the usage of both contextual and parametric knowledge.
This indicates that \MethodName can more accurately extract features related to knowledge selection behaviours through the SAE and use them to steer the generation more effectively.

Second, we find \emph{\MethodName outperforms contrastive decoding methods} DoLa and CAD, especially in steering the usage of parametric knowledge.
Though contrastive decoding strategies can effectively improve the use of contextual knowledge, they struggle to steer the use of parametric knowledge.
In contrast, \MethodName can more effectively steer the usage of both knowledge by adding and removing the desired and undesired functional features, which we will further analyse in the later ablation study.

Moreover, \emph{\MethodName surpasses the non-inference-time controlling method} ICL.
It suggests the \MethodName can both effectively and efficiently control the knowledge selection behaviours of LLMs.
It also suggests a promising capability of representation engineering to control the behaviours of LLMs at inference time in practical applications.

\begin{figure*}[t]
    \centering
    \begin{subfigure}[b]{0.32\textwidth}
        \centering
        \includegraphics[width=\linewidth]{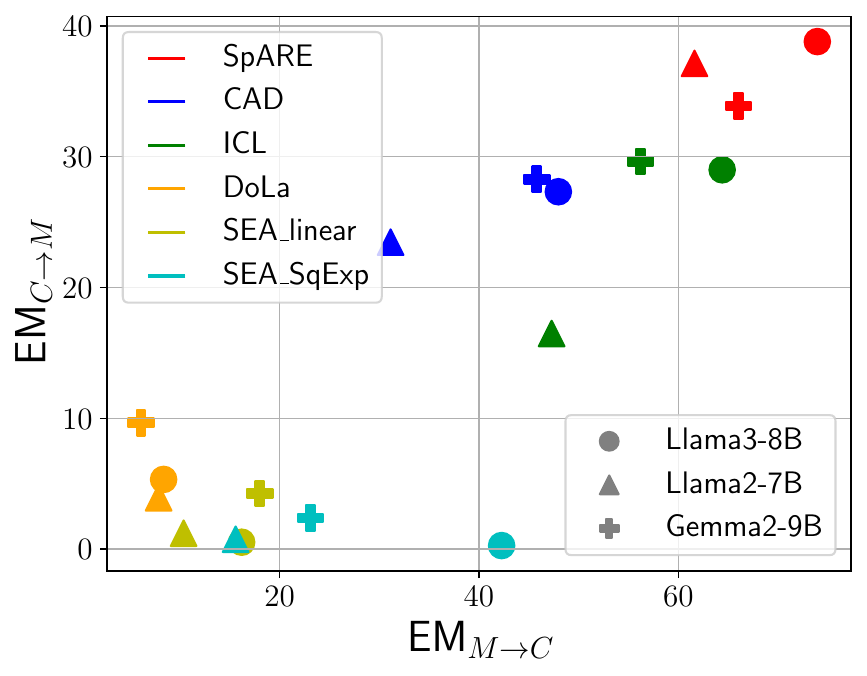}
        \caption{Capability of changing behaviours.}
        \label{fig:change-behaviour}
    \end{subfigure}
    \begin{subfigure}[b]{0.32\textwidth}
        \centering
        \includegraphics[width=\linewidth]{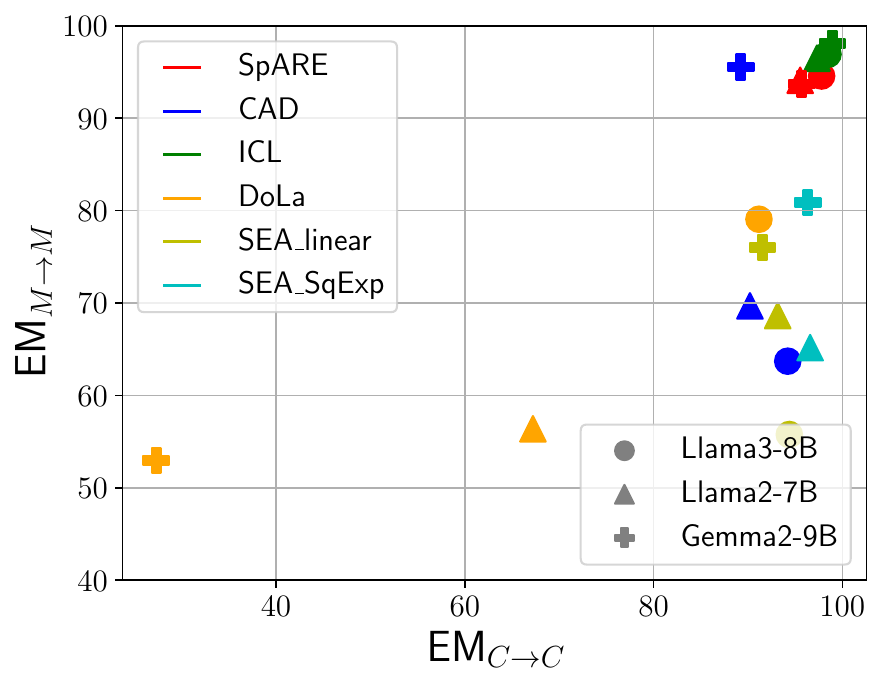}
        \caption{Impact of intervention.}
        \label{fig:mantain-behaviour}
    \end{subfigure}
    \begin{subfigure}[b]{0.32\textwidth}
        \centering
        \includegraphics[width=\linewidth]{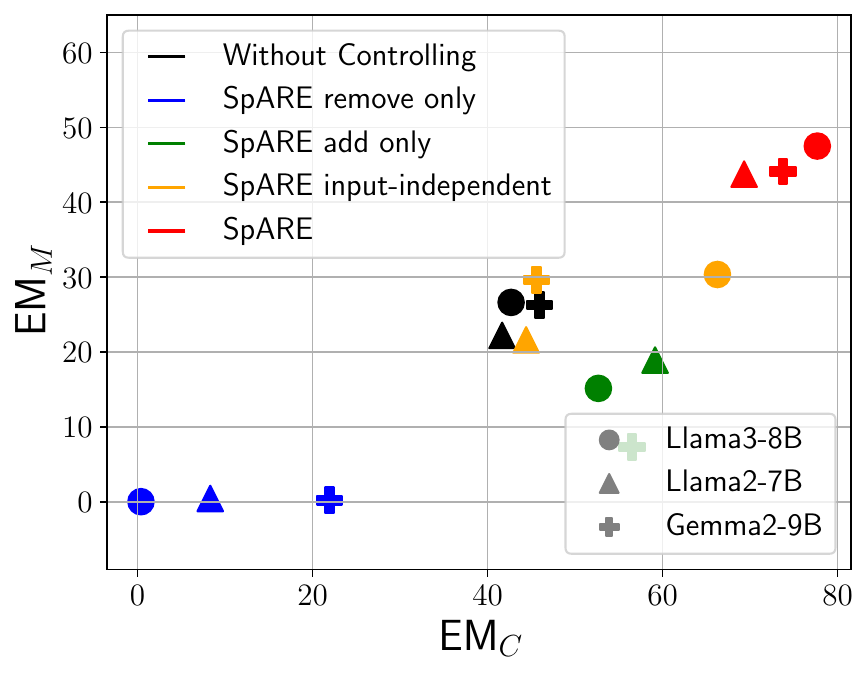}
        \caption{Ablation studies.}
        \label{fig:ablation-study}
    \end{subfigure}
\caption{
Detailed evaluation results of controlling capability on NQSwap.
We use different colours for different methods and use different shapes for different models.
The upper-right area indicates a high performance for all figures.
(a)~presents the capability of changing the behaviour of LLMs, where $x$-axis and $y$-axis are \emctom and \emmtoc, measuring the capability of changing the answer from \ConflictAnswer to \MemorisedAnswer and from \MemorisedAnswer to \ConflictAnswer, respectively;
(b)~presents the capability of maintaining the behaviour when steering to the same behaviour as the original behaviour, where $x$-axis and $y$-axis are \emmtom and \emctoc, measuring the maintaining capability of generating \MemorisedAnswer and \ConflictAnswer, respectively;
(c) present the ablation analysis of \MethodName, $x$-axis and $y$-axis are \emm and \emc.
}
\label{fig:detailed-evaluation}
\end{figure*}

\subsection{Multi-Perspective Controlling Analysis}
In the following, we analyse the controlling capability of \MethodName from different perspectives: \textit{1)} the capability of changing the behaviour (\cref{fig:change-behaviour}), \textit{2)} the potential negative impact of intervention (\cref{fig:mantain-behaviour}), and \textit{3)} the ablation study (\cref{fig:ablation-study}).
\paragraph{Capability of Changing the Behaviours}
Unlike merely comparing overall performance across the entire dataset,
we further examine their capability of changing the original knowledge selection behaviour of LLMs by the following two metrics: 
\begin{description}[leftmargin=*,nosep,topsep=0pt]
\item[\emctom] accuracy of changing the behaviour from generating contextual answers \ConflictAnswer to parametric answers \MemorisedAnswer in the subset of instances where the model generates \ConflictAnswer without controlling.
\item[\emmtoc] accuracy of changing the behaviour from generating parametric answers \MemorisedAnswer to contextual answers \ConflictAnswer in the subset of instances where the model generates \MemorisedAnswer without controlling.
\end{description}

As shown in~\cref{fig:change-behaviour}, \MethodName outperforms contrastive decoding methods and is located in the upper-right area of the figure.
\MethodName also outperforms representation engineering methods. It suggests that the SAE enables accurately extracting features related to knowledge behaviour and thus more effectively changes both the original behaviours of using contextual and parametric knowledge.
We also observe that all methods are less effective in steering toward the use of parametric knowledge than contextual knowledge.
This finding matches the previous works~\citep{conflictbank,conflictqa,competition-of-mechanisms}, which shows LLMs prefer contextual knowledge, and thus more difficult to steer the behaviour of using parametric knowledge.

\paragraph{Impacts of Intervention}
A sufficient but unnecessary intervention can change the behaviour of LLMs, but it also can introduce noise and decrease accuracy.
Here, we investigate the potential negative impact of methods by steering LLMs using the same knowledge they will use without control.
We expect LLMs to maintain their original behaviours, measured by the following two metrics:
\begin{description}[leftmargin=*,nosep,topsep=0pt]
\item[\emmtom] accuracy of maintaining the behaviour of generating \MemorisedAnswer when steering the use of parametric knowledge in the subset of instances where the model generates \MemorisedAnswer without controlling.
\item[\emctoc] accuracy of maintaining the behaviour of generating \ConflictAnswer when steering the use contextual knowledge in the subset of instances where the model generates \ConflictAnswer without controlling.
\end{description}

As shown in \cref{fig:mantain-behaviour}, as it minimally alters the original behaviour when guiding the model to produce similar outcomes.
\MethodName has a close performance to ICL, indicating it can steer the behaviour effectively while introducing a little unnecessary editing.
Though CAD maintains the most accuracy in contextual knowledge, its performance decreases substantially in maintaining the behaviour of generating parametric knowledge.
Finally, while other representation engineering methods may alter the entire model behaviour due to editing of polysemantic features, \MethodName provides a more precise approach to editing through the SAE activations and thus delivers better performance in maintaining the model behaviours.

\paragraph{Ablation Study}
We present the ablation study in the following settings:
\textit{1)} \MethodName input-independent: it uses $\mathbf{z}_C$ and $\mathbf{z}_M$ to steer the generation without calculating $\mathbf{z}^-$ and $\mathbf{z}^+$ based on the input activation;
\textit{2)} \MethodName remove only: it edits the hidden states by only removing the functional features of the undesired behaviour;
\textit{3)} \MethodName add only: it edits the hidden states by only adding the functional features of the desired behaviour.
As shown in \cref{fig:ablation-study}, we can see that every ablation results in a significant controlling capability decrease.
The input-independent editing strategy that omits the calculations of \cref{eq:remove-act} and \cref{eq:add-act} fails to steer the usage of knowledge and obtain results that are close to the ones we obtain without controlling.
The results of \MethodName "remove only" obtain a zero accuracy on both \emm and \emc,  indicating that the model cannot keep the original behaviour and also cannot generate answers toward another behaviour.
This suggests that \MethodName can effectively remove the functional features of the original behaviour.
Moreover, \MethodName "add only" leads to worse performance than without controlling, suggesting the importance of removing the features of undesired knowledge selection behaviour.

\section{Analysis and Discussion}
\label{sec:analysis-and-discussion}
\subsection{Analysing the Layer Choice} \label{sec:edit-layers}
\begin{figure}
	\centering
	\begin{minipage}[t]{\linewidth}
        \centering
    \begin{subfigure}[b]{0.49\linewidth}
        \centering
        \includegraphics[width=\linewidth]{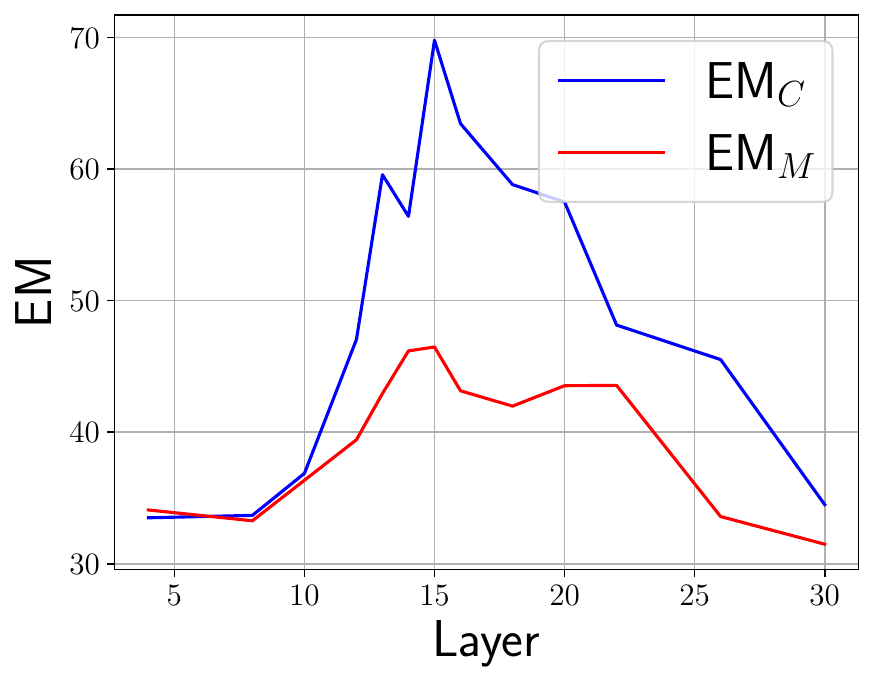}
        \caption{Llama3-8B}
    \end{subfigure}
    \begin{subfigure}[b]{0.49\linewidth}
        \centering
        \includegraphics[width=\linewidth]{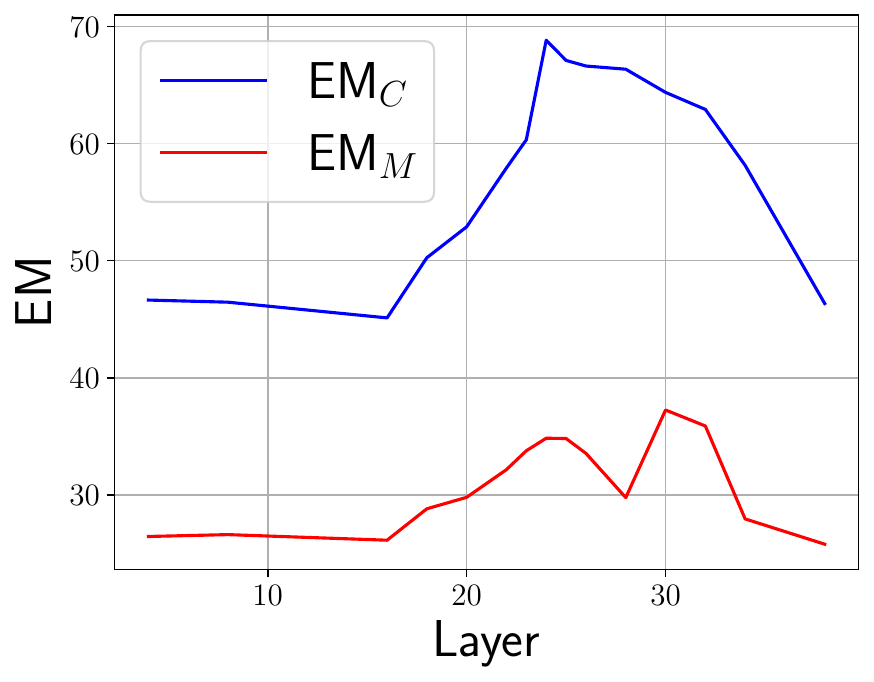}
        \caption{Gemma2-9B}
    \end{subfigure}
    \caption{Effectiveness of \MethodName on editing different layers individually.}
\label{fig:layer-control-effectivenes}
	\end{minipage}
	\\ \vspace{2em}
 
	\begin{minipage}[t]{\linewidth}
        \centering
    \begin{subfigure}[b]{0.49\linewidth}
        \centering
        \includegraphics[width=\linewidth]{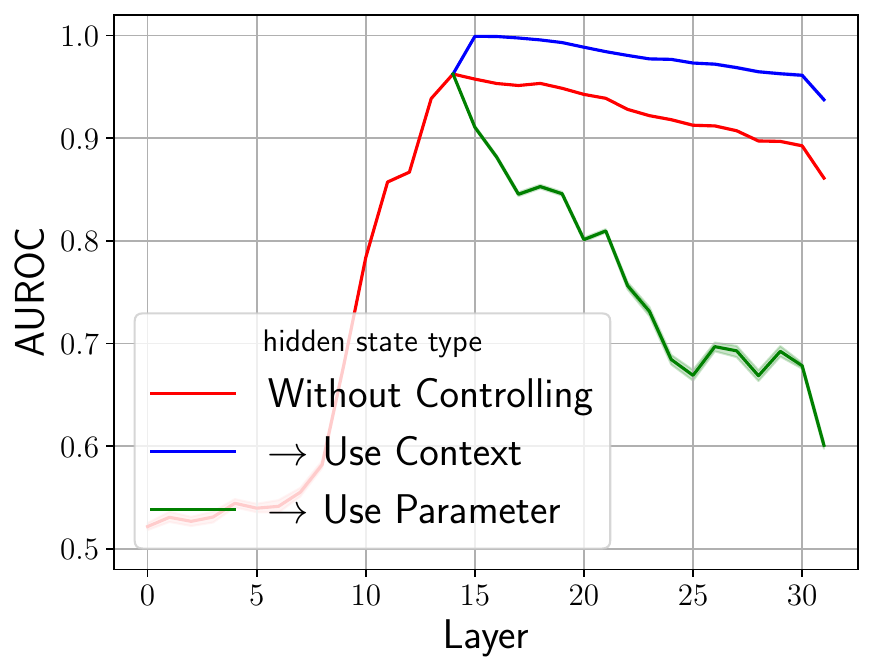}
        \caption{Probing results}
        \label{fig:probing-change}
    \end{subfigure}
    \begin{subfigure}[b]{0.49\linewidth}
        \centering
        \includegraphics[width=\linewidth]{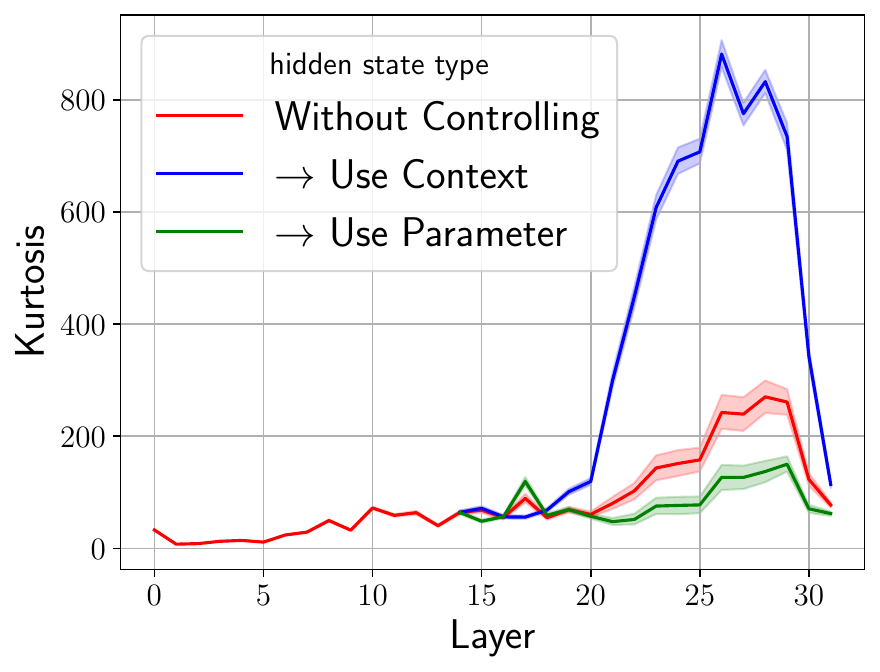}
        \caption{Skewness pattern}
        \label{fig:skewness-change}
    \end{subfigure}
    \caption{The residual stream changes after applying \MethodName to Llama3-8B at the 15th layer.}
\label{fig:residual-stream-changes}
	\end{minipage}
\end{figure}

We present the results of editing multiple layers in~\cref{tab:main-results}; here, we analyse the effectiveness of \MethodName by editing each layer individually.
As shown in \cref{fig:layer-control-effectivenes}, we find \MethodName can control the behaviour of LLMs most effectively at mid-layers for both Llama3-8B and Gemma2-9B.
These layers are also where we can detect knowledge conflict most accurately, as shown in \cref{fig:knowledge-conflict-probing} and \cref{sec:more-knowledge-conflict-detection}.
This supports the practical application of inference-time intervention to control knowledge selection behaviour, where \MethodName can effectively steer the generation once we detect the conflict.

The effectiveness of steering the behaviours in middle layers also matches previous findings~\citep{task-vector,DBLP:conf/acl/PanG0C23,DBLP:conf/iclr/ToddLSMWB24}, that suggest that the middle layers of LLMs contain the functional feature that drives a task.
To the best of our knowledge, we are the first to accurately extract this functional feature using pre-trained SAEs.

\subsection{Analysing the Residual Stream}
We analyse how the residual stream changes after applying \MethodName.
Here, we edit the hidden states from \ConflictEvidenceDataset$=\{(Q, E_C)\}$ at the 15th layer to steer the contextual and parametric knowledge usage. 

In~\cref{fig:probing-change}, we present the probing results on the residual stream using the same probing model described in~\cref{sec:knowledge-conflict-detection}.
We observe that when we steer towards the usage of parametric knowledge, the probing performance decreases \emph{immediately} (green line), indicating that the signal of knowledge conflict fades quickly, and the representations of activations become closer to \MemorisedEvidenceDataset, thus making it more difficult for the probing model to distinguish whether a given activation is from \ConflictEvidenceDataset or \MemorisedEvidenceDataset.
In contrast, when we steer towards using contextual knowledge, the probing performance increases (blue line), indicating the signal of the conflict increases, and the representations of activations become more different from \MemorisedEvidenceDataset, making it easier for the probing model to distinguish whether a given activations is from \ConflictEvidenceDataset or \MemorisedEvidenceDataset.

\begin{figure}
    \begin{subfigure}[b]{0.49\linewidth}
        \centering
        \includegraphics[width=\linewidth]{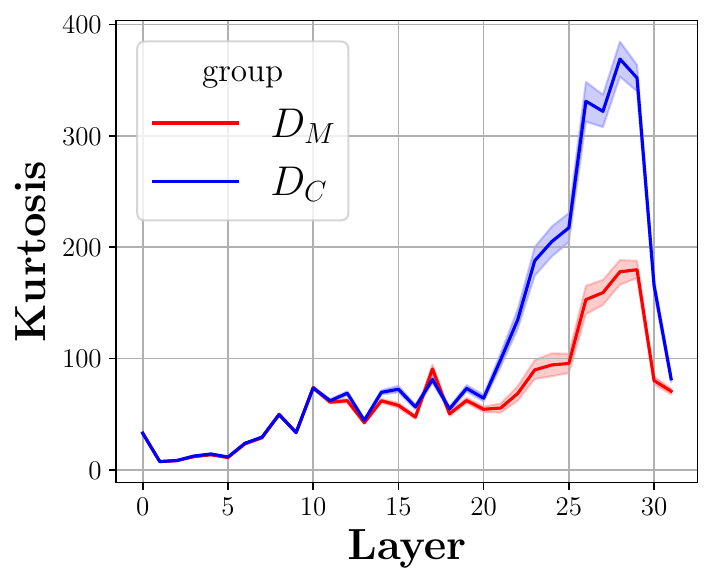}
        \caption{Llama3-8B}
    \end{subfigure}
    \begin{subfigure}[b]{0.49\linewidth}
        \centering
        \includegraphics[width=\linewidth]{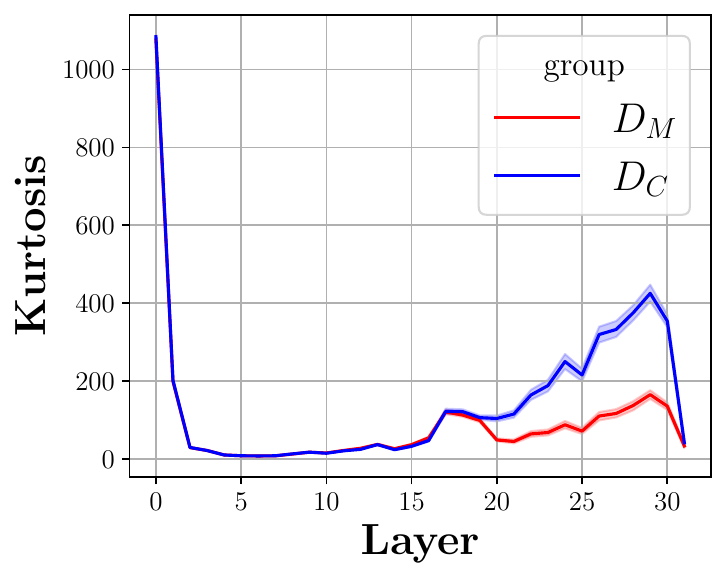}
        \caption{Llama2-7B}
    \end{subfigure}
    \caption{The skewness patterns of the residual steam when LLMs select different sources of knowledge to generate the answer without controlling in NQSwap. 
    }
    \label{sec:analysing-skewness-main-paper}
\end{figure}

In~\cref{fig:skewness-change}, we find the skewness of the representation from the residual stream -- measured by Kurtosis -- shows distinct patterns after applying \MethodName.
We observe that when we apply \MethodName to steer the usage of contextual knowledge at the 15th layer, the residual stream becomes significantly more skewed starting from \emph{later layers}---the 19th layer (blue line); in contrast, when we use parametric knowledge, the residual stream becomes less skewed (green line).
Moreover, in~\cref{sec:analysing-skewness-main-paper}, we analyse the skewness pattern when LLMs freely select knowledge to generate answers without controlling.
We find the residual stream of \UseContextDataset, where the model uses contextual knowledge to generate answers, is significantly more skewed than \UseParameterDataset from the 19th layer.
Thus, the skewness pattern changes shown in~\cref{fig:skewness-change} can indicate that \MethodName steers the knowledge selection behaviours.

We provide more analysis of the representation patterns in~\cref{sec:distribution-pattern} and~\citet{residual-stream-knowledge-conflict}.
In this work, we only provide our empirical observation on the representation pattern and leave investigating the reasons in future works.

\section{Related Works}
\paragraph{Representation Engineering}
Many studies focus on \emph{mechanistic interpretability} to understand the LLMs  by analysing the activities and connections of individual network components, such as circuits~\citep{mathematical-framework,induction-head} and neurons~\citep{feed-forward-layers-are-key-value-memories,rome} 
However, though mechanistic interpretability can successfully explain simple mechanisms, it often struggles with more complex phenomena~ \citep{representation-engineering}.
Differently, \emph{representation engineering} \cite{actadd,spectral-editing,representation-engineering} offers a complementary approach. 
It focuses on the characteristics of representations rather than lower-level mechanisms, providing a framework for understanding complex systems at a higher level of abstraction. It has shown more promise in interpreting higher-level behaviours of LLMs at scale.
Some works modify model activations to change behaviours \citep{DBLP:conf/acl/RavfogelEGTG20,DBLP:conf/acl/IskanderRB23,DBLP:conf/icml/LiuY0Z24,DBLP:journals/corr/abs-2310-01405,DBLP:conf/nips/0002PVPW23}, and some works extract latent vectors and leveraging these vectors to regulate the model's inference \citep{DBLP:journals/corr/abs-2308-10248,DBLP:conf/acl/SubramaniSP22,DBLP:journals/corr/abs-2312-06681}.

\paragraph{Knowledge Conflicts}
\emph{Knowledge conflicts} refer to discrepancies among contextual and parametric knowledge~\citep{DBLP:conf/emnlp/ChenZC22,DBLP:conf/iclr/Xie0CL024}.
\citet{xu2024knowledge} identify three types of knowledge conflicts: \emph{inter-context}~\citep{DBLP:conf/emnlp/ZhangC21,DBLP:conf/aaai/DuBM22,DBLP:conf/ijcnlp/PanCKW23,packing}, \emph{context-memory}~\citep{nqswap,DBLP:conf/iclr/Xie0CL024,minder2024controllable}, and \emph{intra-memory conflicts}~\citep{DBLP:journals/corr/abs-2311-05232}.
In this work, we focus on context-memory knowledge conflict, which refers to conflicts between the contextual knowledge and the parametric knowledge encoded in the model parameters. \citet{competition-of-mechanisms} and \citet{jin2024cutting} investigate the mechanisms of attention heads and feed-forward networks of LLMs when context-memory knowledge conflict occurs.

\paragraph{Sparse Auto-Encoder}
Sparse Auto-Encoders (SAEs) have been introduced as a post-hoc analysis tool to identify disentangled features within uncompressed representations of an LLM~\citep{yun-etal-2021-transformer, bricken2023monosemanticity, cunningham2024sparse}.
SAEs are trained with sparsity regularisation to learn a sparse, overcomplete basis that characterises the activation space of an LLM~\citep{bereska2024mechanistic}.
\citet{marks2024sparse} showed that the features learned by SAEs can identify sparse circuits in LLMs.
\citet{templeton2024scaling} showed the possibility of searching for monosemantic features and steering LLMs' generation. \citet{Chalnev2024ImprovingSV} improves steering vectors using SAEs. \citet{gur2025enhancing} generates description for representations learned through SEAs. \citep{lan2024sparse} finds universal representation across different language models.

\section{Conclusions}
We investigated the context-memory knowledge conflicts in LLMs.
We identify that knowledge conflicts can be detected by probing the residual stream of the model (\cref{sec:detection}) and propose \MethodName (\cref{sec:spare}), a training-free representation engineering method that leverages pre-trained SAEs to effectively and efficiently control the knowledge selection behaviour of LLMs at inference time.
Our experimental results on ODQA tasks show that \MethodName produces more accurate results than existing representation engineering and contrastive decoding methods (\cref{sec:experiments}).
By providing a mechanism to steer knowledge selection behaviours at inference time, \MethodName offers a promising approach to managing knowledge conflicts in LLMs without significant computational overhead.
Additionally, we investigate the residual stream of LLMs under knowledge conflicts (\cref{sec:analysis-and-discussion}). We find that \textit{1)} the knowledge selection behaviour is more steerable at the middle layers of LLMs; \textit{2)} the residual stream shows a significantly more skewed representation when models using contextual knowledge compared to using parametric knowledge.

\section*{Limitations}
While our proposed method, \MethodName, demonstrates effective control over knowledge selection behaviours in LLMs, there are several limitations to consider.
First, the approach relies on pre-trained SAEs to identify and manipulate functional features within the internal activations of the model, so it may not directly apply to models where pre-trained SAEs are unavailable or cannot be efficiently trained.
Second, our experiments are conducted on specific ODQA tasks involving context-memory knowledge conflicts.
While the results are promising in this setting, it is unclear how well the method generalises to other types of tasks or conflicts, such as those involving complex reasoning, multi-hop questions, or long-form generation.
Finally, the control over knowledge selection behaviours is evaluated primarily in terms of steering the model towards either contextual or parametric knowledge.
In practice, the decision about which knowledge source to trust may not be binary or require a more elaborate approach, such as using another model as a critic~\citep{macnoise}.

\section*{Acknowledgments}
Yu Zhao and Xiaotang Du were partly supported by the UKRI Centre for Doctoral Training in Natural Language Processing, funded by UK Research and Innovation (grant EP/S022481/1) and the University of Edinburgh, School of Informatics.
Alessio Devoto was supported by Sapienza Grant RM1221816BD028D6 (DeSMOS).
Giwon Hong was supported by the ILCC PhD program (School of Informatics Funding Package) at the University of Edinburgh, School of Informatics.
Aryo Pradipta Gema was supported by the United Kingdom Research and Innovation (grant EP/S02431X/1), UKRI Centre for Doctoral Training in Biomedical AI at the University of Edinburgh, School of Informatics.
Xuanli He was funded by an industry grant from Cisco.
Pasquale Minervini was partially funded by ELIAI (The Edinburgh Laboratory for Integrated Artificial Intelligence), EPSRC (grant no.\ EP/W002876/1), an industry grant from Cisco, and a donation from Accenture LLP.
This work was supported by the Edinburgh International Data Facility (EIDF) and the Data-Driven Innovation Programme at the University of Edinburgh.

\bibliography{references}

\clearpage

\appendix

\section{More Analysis of Knowledge Conflict Probing}
\label{sec:more-knowledge-conflict-detection}

\subsection{Details of Probing Model}
We train the probing model with an L$_{1}$ norm regularisation for all probing experiments. The training objective is $\mathcal{L}=-\log P(y=y_i) + \lambda \Vert W \Vert_1$, where we set $\lambda$ to $3\times10^{-4}$ and $y_i$ is the label. We train $20$ times with different random seeds for each probing task, and we report the average and deviation in our experiments.
We split the training and test datasets for the probing tasks, ensuring no overlapping questions between them.
\subsection{More Probing Results}
\begin{figure*}[t]
    \centering
    \begin{subfigure}[b]{0.32\textwidth}
        \centering
        \includegraphics[width=\linewidth]{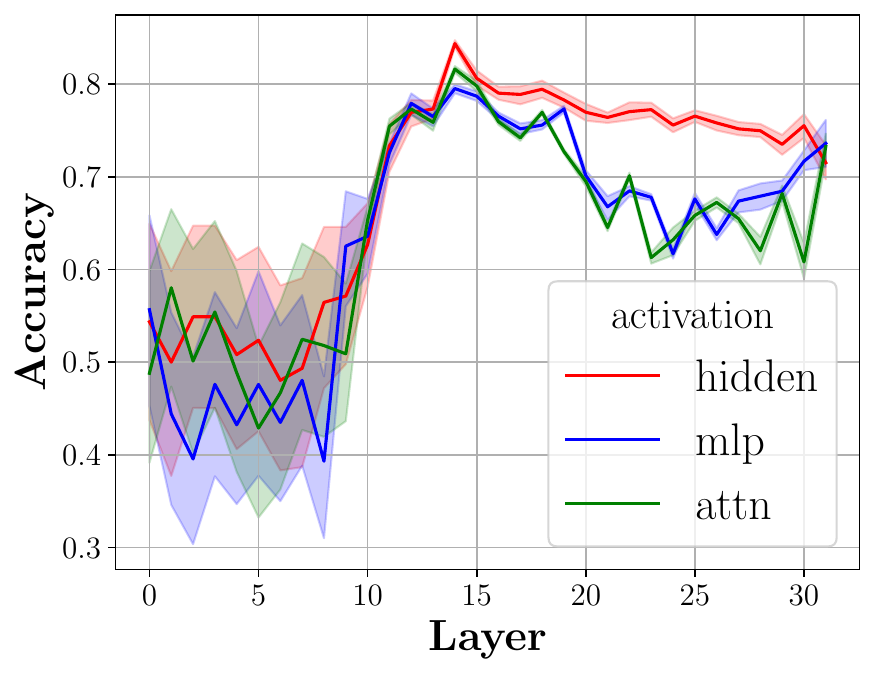}
    \end{subfigure}
    \begin{subfigure}[b]{0.32\textwidth}
        \centering
        \includegraphics[width=\linewidth]{MINT-figures/KC-detect/Llama-2-7b-hf_nqswap_AUROC.pdf}
    \end{subfigure}
    \begin{subfigure}[b]{0.32\textwidth}
        \centering
        \includegraphics[width=\linewidth]{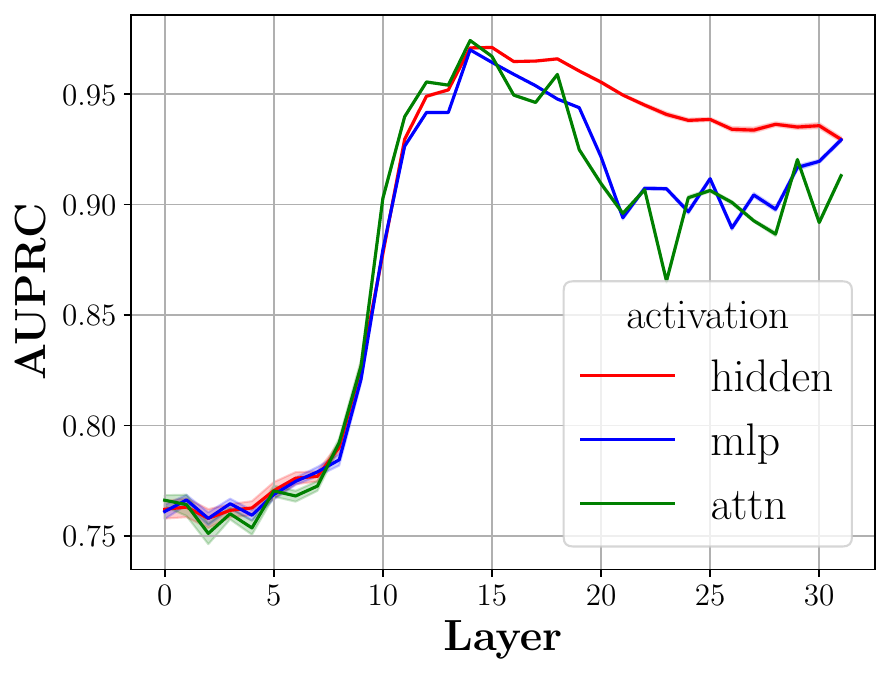}
    \end{subfigure}
\caption{Knowledge conflict probing results using Llama2-7B on NQSwap.}
\label{fig:knowledge-conflict-probing-llama2-nqswap}
\end{figure*}

\begin{figure*}[t]
    \centering
    \begin{subfigure}[b]{0.32\textwidth}
        \centering
        \includegraphics[width=\linewidth]{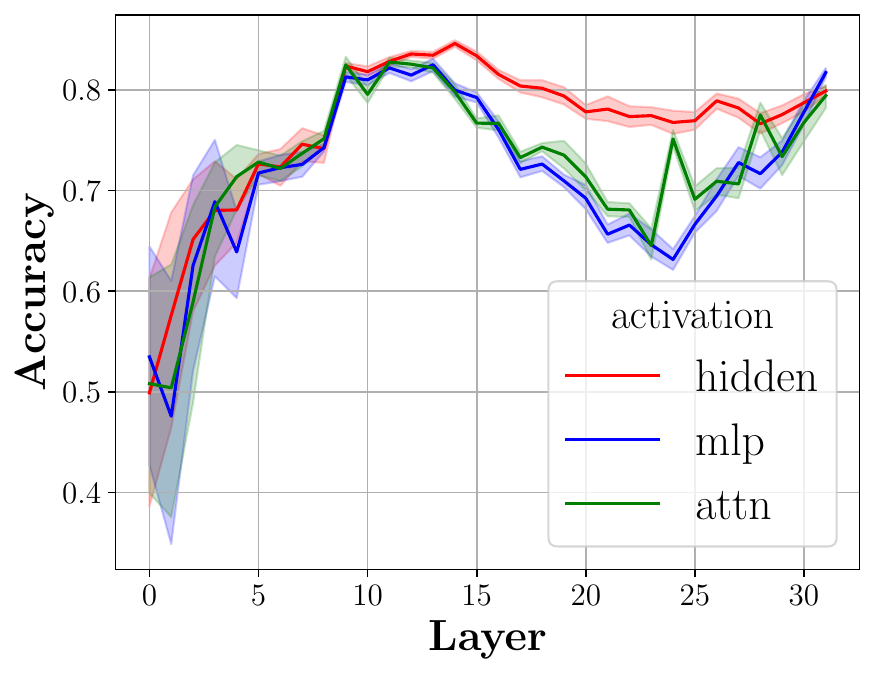}
    \end{subfigure}
    \begin{subfigure}[b]{0.32\textwidth}
        \centering
        \includegraphics[width=\linewidth]{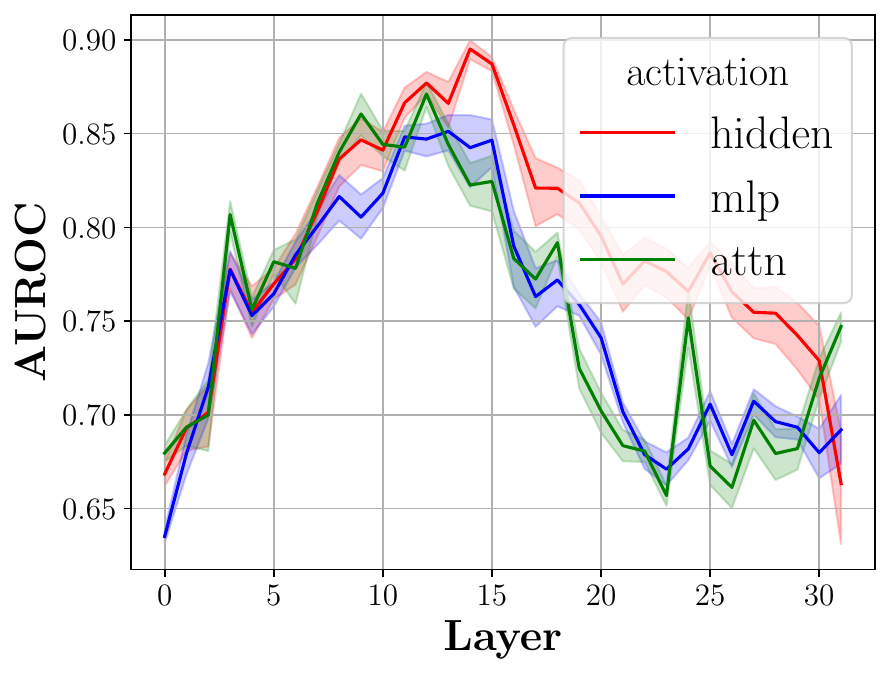}
    \end{subfigure}
    \begin{subfigure}[b]{0.32\textwidth}
        \centering
        \includegraphics[width=\linewidth]{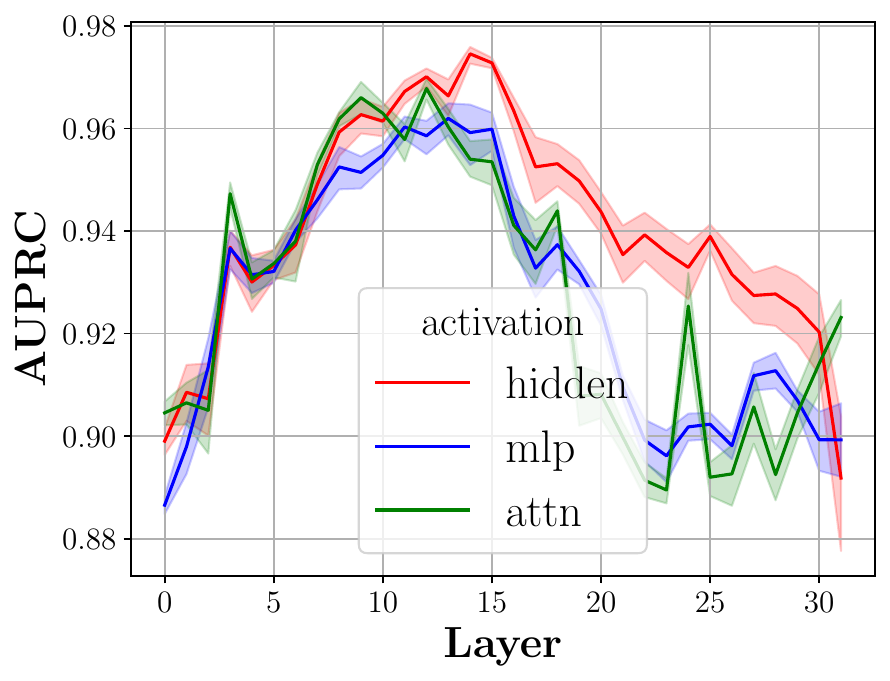}
    \end{subfigure}
\caption{Knowledge conflict probing results using Llama2-7B on Macnoise.}
\label{fig:knowledge-conflict-probing-llama2-macnoise}
\end{figure*}

We present the knowledge conflict probing results of Llama2-7B on NQSwap and Macnoise on \cref{fig:knowledge-conflict-probing-llama2-nqswap} and \cref{fig:knowledge-conflict-probing-llama2-macnoise} using accuracy, AUROC and AUPRC as metrics.
We provide more analysis of probing the residual stream under knowledge conflict in our preliminary study~\citep{residual-stream-knowledge-conflict}.

\section{Sparse Auto-Encoders Details}
\label{sec:details-of-sae}
Both the SAEs of Llama3-8B from ElethuerAI and Llama2-7B pre-trained by us have a latent dimension $n=131072$, $32$ times larger than the number of dimensions of the hidden state size $d=4096$, and use ReLU as activation function $\sigma$ (\cref{eq:sae}). Gemma2-9B uses JumpReLU~\citep{jumprelu} as the activation function. GemmaScope~\citep{gemma-scope} provides different sizes of SAEs, and we use the size of $n=131072$ in the experiment.
Following~\citet{scaling-and-evaluating-sae}, we pre-train the SAEs for Llama2-7B with TopK activation functions: $\mathbf{z} = \text{TopK} \left( \mathbf{W}_\theta(\mathbf{h} - \mathbf{b}) \right)$, which only keeps the $k$ largest latents during pre-training, while does not use sparsity regularisation during pre-training.
The loss is calculated by the L$_{2}$ norm of reconstruction error: $\mathcal{L} = \Vert \mathbf{h} - \hat{\mathbf{h}} \Vert_2^2$.
We pre-train SAE models\footnote{\url{https://github.com/EleutherAI/sae}} with $10$B tokens sampled from RedPajama\footnote{\url{https://huggingface.co/datasets/togethercomputer/RedPajama-Data-V2}} and use the hyperparamters determined by~\citet{scaling-and-evaluating-sae}.
The pre-training for an SAE for a certain layer of hidden states costs about $300$ 80G A100 GPU hours.\footnote{We provide our pre-trained SAEs in \url{https://huggingface.co/yuzhaouoe/Llama2-7b-SAE/tree/main}}
Though it costs non-trivial resources for pre-training SAEs, we believe it is valuable to explore using SAEs to resolve knowledge conflicts for the following reasons:
\textit{1)} SAEs are general models for interpreting the representation of LLMs, which have broader applications beyond steering knowledge selection behaviours;
\textit{2)} SAEs are becoming popular tools for interpreting LLMs with rising numbers of open-resource frameworks and pre-trained models released recently, so we can use the pre-trained SAEs conveniently rather than pre-training SAEs by ourselves in the future.

\section{Implementation Details}
\label{sec:implementation-details}

\subsection{Collecting Activations}
\label{sec:collecting-activation-details}
Collecting activations is an essential step in representation engineering methods~\citep{representation-engineering,spectral-editing}, where we will extract desired features from the collected activations and use these features to edit the activations of LLMs.
In \MethodName, we first sample demonstrations from the development set using $5$ seeds.
Then, we use these demonstrations to test the rest of the questions with conflict evidence \ConflictEvidence.
Based on the predictions, we split the instance into \UseContextDataset and \UseParameterDataset and collect their corresponding hidden states, denoting as $\{\mathbf{h}^j_C\}_{j=1}^N$ and $\{\mathbf{h}^j_M\}_{j=1}^N$. 
Then, the corresponding layer's SAE encodes them to  $\{\mathbf{z}_{C}^j\}_{j=1}^N$ and $\{\mathbf{z}_{M}^j\}_{j=1}^N$.
Here, we use $j$ to index the collected instances and use $i$ to index the SAE's activation.
One strategy to highlight the values of behaviour-related activation is weighted averaging $\{\mathbf{z}_{C}^j\}_{j=1}^N$ and $\{\mathbf{z}_{M}^j\}_{j=1}^N$ by the confidence of generating a specific answer.
Specifically, denote the confidence of generating answers \ConflictAnswer and \MemorisedAnswer conditioned on $\mathbf{h}^j_C$ and $\mathbf{h}^j_M$ as 
\begin{align*}
    \lambda^j_C & = \frac{\log P(C \mid \mathbf{h}^j_C)}{\log P(C \mid \mathbf{h}^j_C) + \log P(M \mid \mathbf{h}^j_C)} \\
    \mu^j_M & = \frac{\log P(M \mid \mathbf{h}^j_M)}{\log P(C \mid \mathbf{h}^j_M) + \log P(M \mid \mathbf{h}^j_M)},
\end{align*}
where $P(\cdot \mid \mathbf{h})$ is calculated by the output of LLMs.
We use the normalised confidence $\lambda^{'j}_C$ and $\mu^{'j}_M$ as weight to average $\{\mathbf{z}_{C}^j\}_{j=1}^N$ and $\{\mathbf{z}_{M}^j\}_{j=1}^N$, respectively: 
\begin{equation*}
    \overline{\mathbf{z}_C} = \sum_{j=1}^{N} \lambda^{'j}_C \mathbf{z}_C^j, \ \text{ and }\   \overline{\mathbf{z}_M} = \sum_{j=1}^{N}  \mu^{'j}_M \mathbf{z}_M^j.
\end{equation*}
We find this strategy brings a slight improvement compared to directly average $\{\mathbf{z}_{C}^j\}_{j=1}^N$ and $\{\mathbf{z}_{M}^j\}_{j=1}^N$.

\subsection{Identifying Functional Activations}
\label{sec:detail-indentify-functional-activations}
We identify the activations that steer the usage of contextual and parametric knowledge by calculating the mutual information between activation $Z_i$ and the generated answer $Y=\{C, M\}$:
\begin{equation*}
    I(Z_i; Y) = \sum_{z_{i}\in Z_i} \sum_{y\in\{C, M\}}P(z_{i}, y) \log \frac{P(z_{i}, y)}{P(z_{i})P(y)}.
\end{equation*}
Since a higher $I(Z_i;Y)$\footnote{ We estimate mutual information between continuous variables $Z_i$ and discrete labels $Y$ using \url{https://scikit-learn.org/1.5/modules/generated/sklearn.feature_selection.mutual_info_classif.html}} indicates a higher dependency between $Z_i$ and the knowledge selection behaviour, we sorted $\{I(Z_i;Y)\}_{i=1}^n$ in descending order, and consider the top $k$ activation work as functional activations.
To decide the number of selected activations $k$, we introduce a hyperparameter $K$. We then choose $k$ such that:
\begin{equation}
\label{eq:topKactivation}
k=\argmin_k \sum_{i=1}^k\frac{I(Z_i;Y)}{\sum_{j=1}^nI(Z_j;Y)} \geq K,    
\end{equation}
which means selecting the top $k$ SAE activations with the proportion $K\%$ in the sum of all mutual information $\sum_{j=1}^nI(Z_j;Y)$.
One potential improvement is normalising mutual information based on entropy, which has shown better properties for comparing the importance of features.
We determine the top activations in each layer individually rather than ranking all mutual information across layers.

The proportion $K$ abstracts the exact number of activations to select.
We expect the same types of SAEs, e.g., pre-trained by TopK~\citep{scaling-and-evaluating-sae} or JumpReLU~\citep{jumprelu}, can share similar values of $K$ for controlling.
We present the relation between $k$ and $K$ in~\cref{sec:mutual-information-distribution}.
We also present the selected Gemma2-9B SAEs activations used by \MethodName in~\cref{sec:gemma-selected-activations}.

\subsection{Impact of the Size of the Collected Activations}
\label{sec:hidden-num}
\begin{figure}[t]
        \centering
    \begin{subfigure}[b]{0.49\linewidth}
        \centering
        \includegraphics[width=\linewidth]{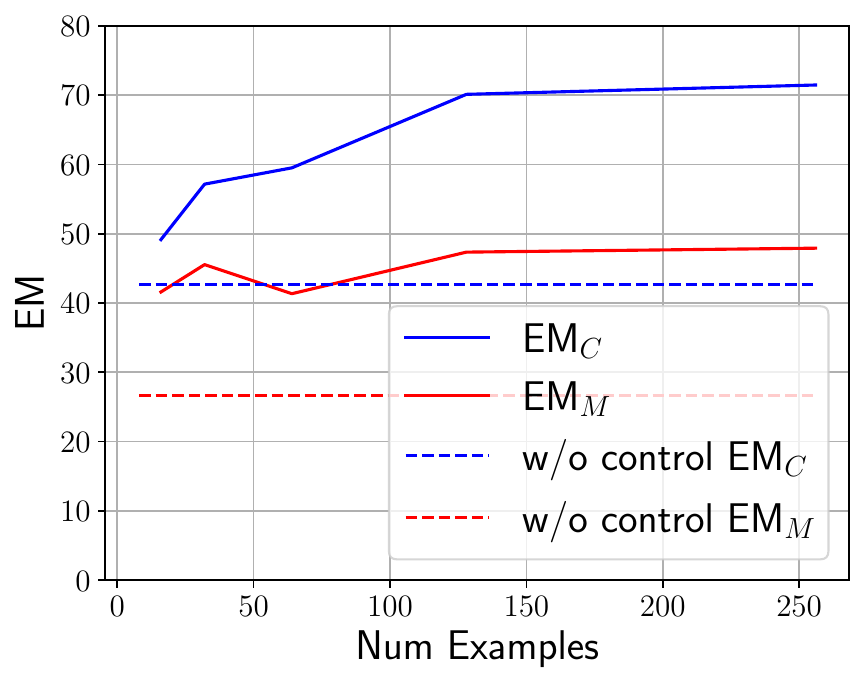}
        \caption{Use different numbers of activations to calculate mutual information. (\cref{sec:measure-correlation})}
        \label{fig:hidden-num-mi}
    \end{subfigure}
    \begin{subfigure}[b]{0.49\linewidth}
        \centering
        \includegraphics[width=\linewidth]{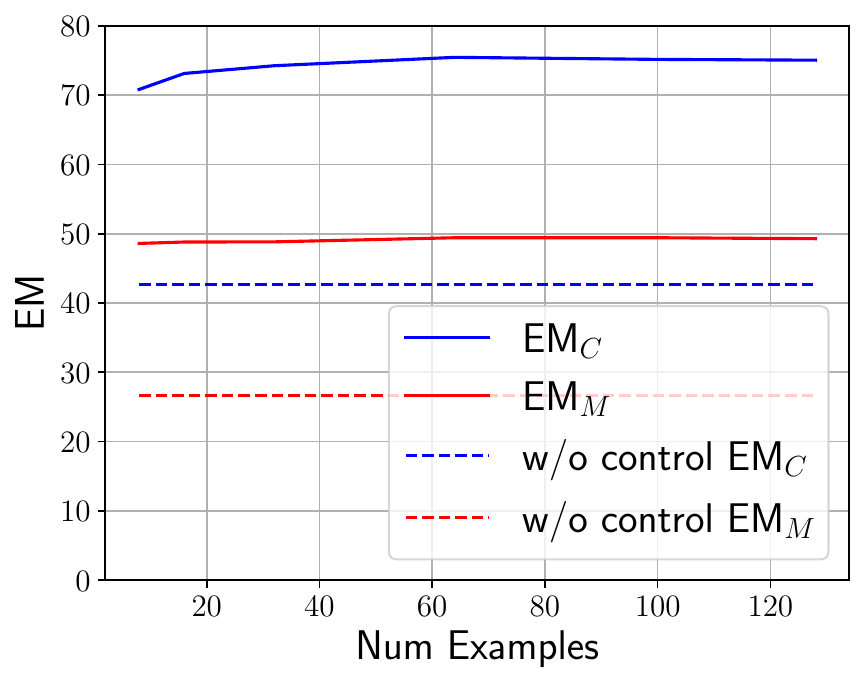}
        \caption{Use different numbers of activations to calculate $\mathbf{z}_C$ and $\mathbf{z}_M$. (\cref{sec:prepare-activation})}
        \label{fig:hidden-num-z}
    \end{subfigure}
    \caption{The impact of the number of the collected hidden states $N$ on the controlling performance.}
\label{fig:num-hidden}
\end{figure}

We collect the hidden states $\{\mathbf{h}_C^j\}_{j=1}^N$ and $\{\mathbf{h}_M^j\}_{j=1}^N$ to calculate the value of functional activations $\mathbf{z}_C$ and $\mathbf{z}_M$, and we also use  $\{\mathbf{z}_C^j\}_{j=1}^N$ and $\{\mathbf{z}_M^j\}_{j=1}^N$ to estimate the mutual information.
Here, we analyse the impact of $N$ on the controlling performance.
As shown in~\cref{fig:hidden-num-mi}, the performance increases when we use $8$ examples to $128$ for calculating the mutual information, while collecting more activations brings slight improvement.
In~\cref{fig:hidden-num-z}, the performance does not increase until using $64$ examples to calculate $\mathbf{z}_C$ and $\mathbf{z}_M$.
The above analysis indicates that \MethodName needs at least $128$ activations to achieve a high controlling capability.

\subsection{Development Set and Demonstrations}
\label{sec:development-set-and-demonstrations}
We held out a demonstration set consisting of $128$ instances from each dataset for each model.
Each question in the demonstration set can be answered correctly by the corresponding model without providing evidence; more specifically, we test it with the close-book setting with few-shot examples to align the answer format.
We use the non-conflict context \MemorisedEvidence and memorised answer \MemorisedAnswer in demonstrations.
Since the contextual and parametric knowledge are consistent in the demonstration set, they do not provide information about how to resolve the knowledge conflict for the test examples when presented with conflict evidence \ConflictEvidence.
We sample $5$ different sets of demonstrations using $5$ different seeds and present the average results with deviations.
All methods we compared use the same sets of demonstrations for each model.

The held-out demonstration set is also used as the development set to search hyperparameters for each model in each dataset.
We sample a set of demonstrations to align the answer format and use the rest of the instances to evaluate the performance.

\subsection{Implementation Details of Representation Engineering Baselines}
\label{sec:re-baseline-details}

For all representation engineering baselines  TaskVec \citep{task-vector}, ActAdd~\citep{actadd} and SEA~\citep{spectral-editing}, we experimented with performing the intervention at different layers and reported the best performance in each case.
We present the implementation details as follows.

\paragraph{TaskVec}\citep{task-vector}: We first collect the task vectors using a few-shot setup where the few-shot examples contain the conflict evidence \ConflictEvidence along with the memorised answer \MemorisedAnswer. 
We then follow the procedure illustrated in the original paper and extract the Task Vectors as the hidden representation of the last token (\verb|:| in our case) at all layers. 
More specifically, we use the hidden representation of samples that generate answers following the context to create $\overline{\mathbf{h}_C}$, while we use those that follow the parametric knowledge and ignore the context to create $\overline{\mathbf{h}_M}$. 
At inference time, we replace the residual stream at layer $L$ with $\overline{\mathbf{h}_C}$ or $\overline{\mathbf{h}_M}$ to steer the model to follow the context or the parametric knowledge, respectively.    
\paragraph{ActAdd}\citep{actadd}: We collect the activations for ActAdd following the same procedure. For the inference-time intervention, we edit the residual stream by adding  $\overline{\mathbf{h}_C}$ and subtracting $\overline{\mathbf{h}_M}$ to steer the model towards context knowledge, while we perform the opposite to steer the model towards parametric knowledge.
\paragraph{SEA}\citep{spectral-editing}: SEA adopts a different strategy and uses positive, negative and neutral model generations to compute steering vectors.
We consider as neutral the generations that result from a few-shot setup where the demonstrations contain the memorised evidence \MemorisedEvidence and memorised answer \MemorisedAnswer, irrespective of the generated output. 
We then collect activations $\overline{\mathbf{h}_M}$ and $\overline{\mathbf{h}_C}$ following the same procedure illustrated above and use the method described in \citet{spectral-editing} to compute the steering vectors, assuming $\overline{\mathbf{h}_M}$ and $\overline{\mathbf{h}_C}$ encode the positive (follow parametric knowledge) and negative (follow context knowledge) behaviours respectively. 

\subsection{Searching Hyperparameters}
\label{sec:baseline-details}
We search hyperparameters of all methods using the same set of development sets.
We choose hyperparameters based on the \emctom and \emmtoc, which measure the capability of changing the behaviours of LLMs.

\paragraph{DoLa}\citep{dola}: We search the best coefficient $\alpha$ that is used to compare premature and mature logits, evaluating with the "higher-layer" and "lower-layer" settings:
\begin{equation*}
    \log P(y) = \log P_{\text{mature}}(y) - \alpha \log P_{\text{premature}}(y).
\end{equation*}
We test the $\alpha$ ranging from $-10$ to $10$ with an interval of $0.5$.
Finally, we set $\alpha=6.0$ and $\alpha=-8.0$ to steer the model to use contextual and parametric knowledge for Llama2-7B and Llama3-8B; set $\alpha=1.0$ and $\alpha=-1.0$ for Gemma2-9B.
However, based on our experiments, though DoLa has a certain ability to change the behaviours of Gemma2-9B, we do not find a suitable $\alpha$ on both "high-layer" and "low-layer" choices for improving Gemma2-9B on the overall performance measured by \emm and \emc.

\paragraph{CAD}\citep{cad}: We search the best coefficient $\alpha$ that is used to combine the logits with context and the logits without the context:
\begin{equation*}
    \log P(y) = (1+ \alpha) \log P(y \mid c, x) - \alpha \log P(y \mid x).
\end{equation*}
We test the $\alpha$ ranging from $-1.5$ to $1.5$ with an interval of $0.1$.
Finally, we set $\alpha=0.6$ and $\alpha=-0.8$ to steer the model to use contextual and parametric knowledge for Llama2-7B and Llama3-8B; set $\alpha=0.3$ and $\alpha=-0.3$ for Gemma-2-9B.

\paragraph{\MethodName}(Ours): Since no previous work used SAEs to control the knowledge selection behaviour of LLMs, we need first to identify suitable magnitudes of the hyperparameters and then search them in a smaller range in the development set.
We fix the $\alpha=1$ in~\cref{eq:hidden-edit} and test the proportion $K$ to select the activations ranging from $0.1$ to $0.01$ with an interval of $0.01$.
Then, we choose $K=0.07$ and test $\alpha$ ranging from $1.0$ to $3.0$ with an interval of $0.2$.
Finally, we select $K=0.07$ and $\alpha=2$ for Llama3-8B, $K=0.06$ and $\alpha=2.2$ for Llama2-7B.
In our main experiments, \MethodName edits the 13th to the 16th layers of Llama3-8B and the 12th to the 15th layers of Llama2-7B. We do not try other choices because there is no public SAE for Llama2-7B.

Due to the Gemma2-7B's SAEs~\citep{gemma-scope,jumprelu} using different training strategies and activation functions, they show a much more sparse pattern and have different suitable hyperparameters. 
Here, we select $K=0.01$ and $\alpha=3$ to steer contextual knowledge, and $K=0.01$ and $\alpha=1.8$ to steer parametric knowledge.
In our main experiments, \MethodName edits 23, 24, 25, 29, 30 and 31 layers of Gemma2-9B.
We also present the selected top $k$ activations in~\cref{sec:gemma-selected-activations} for further analysis.

\section{Selected SAEs Activations of Gemma2-9B}
\label{sec:gemma-selected-activations}
In~\cref{tab:selected-activations}, we present the selected SAE activations used by \MethodName for steering the knowledge selection behaviour.
We can further interpret them in GemmaScope\footnote{\url{https://www.neuronpedia.org/gemma-scope}}~\citep{gemma-scope}.

\begin{table*}[t]
\centering
\small
\setlength\tabcolsep{10pt}
\begin{tabular}{m{0.5cm}m{5.5cm}m{7.5cm}}
\toprule
 Layer & Use Parametric Knowledge & Use Contextual Knowledge \\  \midrule
23 & 116391,  36331,  85142,   2795,  99547,  63615,  25635, 123378, 105328,
         24132, 113025,  83008,  37706,  60782,  36046, 110864, 101469,  29902,
        129485, 112858, 104185,  17911,   6673,  72533, 108414,  32967,  19761,
        118260, 109917,  55083,  41965,  91874,  74605,  19726, 115338,  80100,
          3042,  48088,  61830,    895,  49288, 120379, 105552,  84782,  14129 & 59646,  66244, 130943, 100165, 103568,  82090, 116937, 108558,  78302,
        100628,  53091,  90600, 124049,  63656, 118525, 119623,  34458, 119574,
         38170,  66293,  14026,  28797, 125520,  76467,  29583,  89951,  32901,
         52256, 130987,  36816,  59062,  58505, 123631,  60183,  11432,  86969,
         11755,  71200,  53746,     33,  57883,  67097, 108617, 112319,   1380,
         47638,  42621,  16859, 130470,   6475, 112033, 101316,  40945,  82574,
         58929,  79660,  81043,  18549,   4537, 130935, 127945,  78809\\ \midrule
24 &10649,  68997,  80242,  38885,  33450,  29004,  34725,  55203,  41474,
         90933, 118013,  76436,   2795,  53138,  41501,  65408, 116855,  12056 & 76071,  55422,  82954,  40832,  68001,  88619, 120959,  92931,  38262,
         83042,  42129,  21413,  74005,  73350,  57270,   6859,  83385,   9263,
          8609,  22968,   8307,  99263,   2415,  59807,  87788,  92845,  88733,
        124321,  25758, 111976,  84892, 104309,  61391,  60162, 128726,  28753,
         62671,  80398,  40150,  28432,  81514,   9463\\ \midrule
25 & 117145,  66103,  55992,   1609, 101788,  28707,  64494,  63602,  81174,
         73438,  16428,   2054,  44642,  12418, 105769,  37692,  33693,  22786 & 77008,  39999,  65977,   3002,  82187, 113845,  35985,  16341, 121937,
         13762,   9468,  70433,  42102,  85578,   3118,  99639,  41828,  58588,
        103815,  70243,  67915, 125985, 113290, 127536,  84912,   2473,  46174,
        100026,  37216,  27820,  81800,  13540, 125213,  79326,  55733,  32460,
         46612\\ \midrule
29 & 70665,  84563,  63717,  45653, 122282,   5001,  67756,  52905, 118450,
         84589,  16721, 119640,  47070,  15218, 117432, 110719,  98957,  11667,
         20824,  31422, 119807,  22664,  81261, 116958 & 65636, 113411,  88779,  19501,  46209,   8584,  71156,  79159,  94888,
           144,  60280,    413, 103986,  74324,  52419,  70057,  30294,  13647,
         37430,  71657, 118541,  12744,  74953, 115544,  19086, 102886,  49216,
         95333,  26177,  89774,  71927,  70989,  23760\\ \midrule
30 & 116964,  47548,  20615,  48375, 128786,   1308,  40865,  22211,  15816,
        107813,  50419, 113319,  97588,  30688, 110627,  56882, 117785,  63602,
         39609,  52155,  99243,  36852, 121514,  73310,    850,  96578 & 84358, 115174,  11363,  28696, 110664,   2831,  24365, 128820,  35092,
         92968,  78722,  22739, 128047, 127030,  77294,  76467,  74131,  56766,
         94697,  58000,  32812,  46910,  82749, 106077,  59596, 103936,   4505,
        129363, 126847,  42463, 120310 \\ \midrule
31 & 61476, 5054,   1364,  18335,  63832,  88313,  35780, 130003,  25371,
        125651,  11685,  24947,   2260,  70799,  92415,  47791,  99787,  88517,
         85499,  75095, 114075, 125055, 109519, 116785, 100449,  37567,  88965,
         59674,  14203, 125588,  70706,  18151 & 121514,  35148,  15479,  65369,  18623,  98225,  52746,  45804, 107893,
         10202,  69463,  83810,  12131, 111417, 115174, 107085,  26328,  75203,
         37430, 127639,  18114,  80704,  68360,  33142,  51607,  96802,  24949,
         97568,  82042,  50826, 110615, 110929,  97833\\ 
\bottomrule
\end{tabular}
\caption{Selected Gemma2-9B SAEs activations with $K=0.01$ (\cref{eq:topKactivation}). "Use Parametric Knowledge" means these activations are positively correlated with the behaviour of selecting parametric knowledge, determined according to our method described in~\cref{sec:measure-correlation}.}
\label{tab:selected-activations}
\end{table*}

\newpage
\begin{figure*}[t]
    \centering
    \begin{subfigure}[b]{0.32\textwidth}
        \centering
        \includegraphics[width=\linewidth]{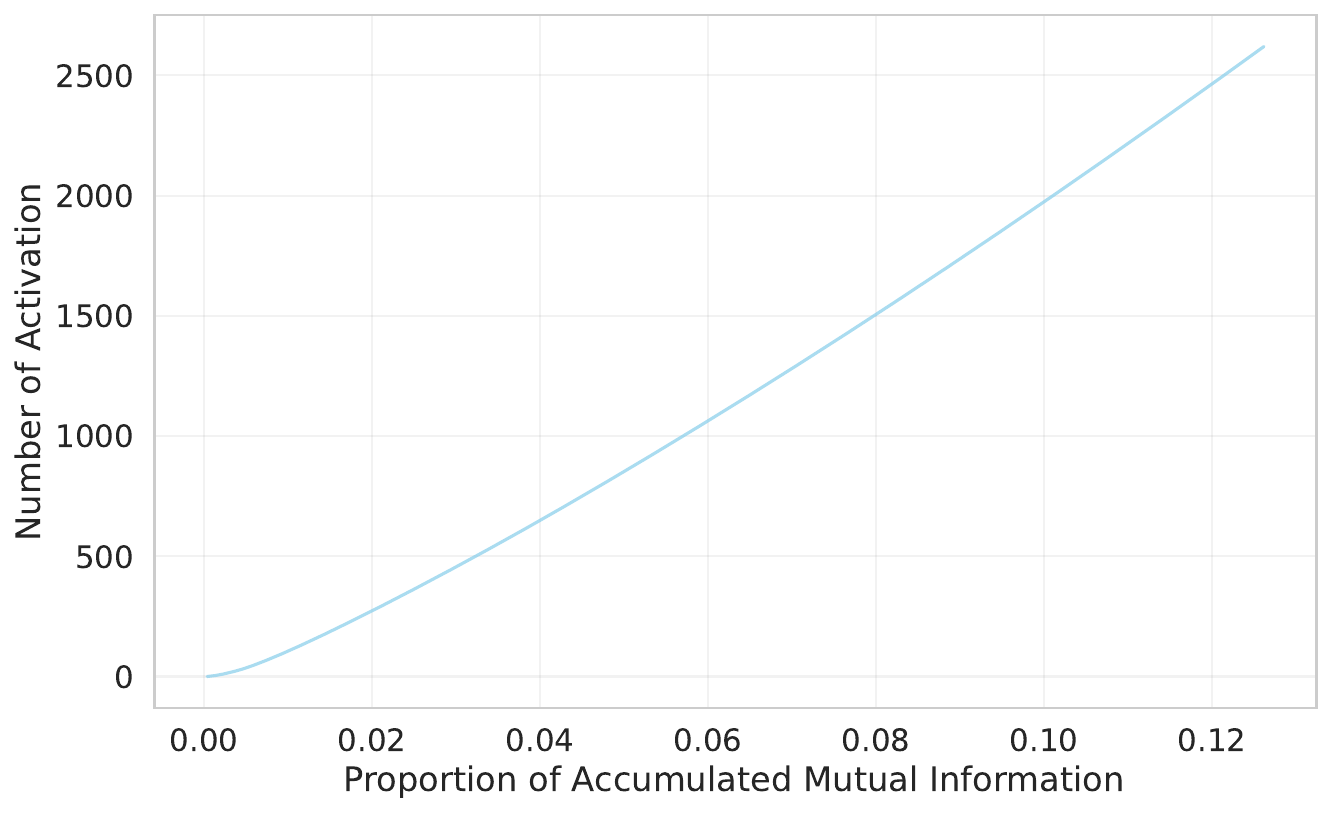}
        \caption{Layer 23}
    \end{subfigure}
    \begin{subfigure}[b]{0.32\textwidth}
        \centering
        \includegraphics[width=\linewidth]{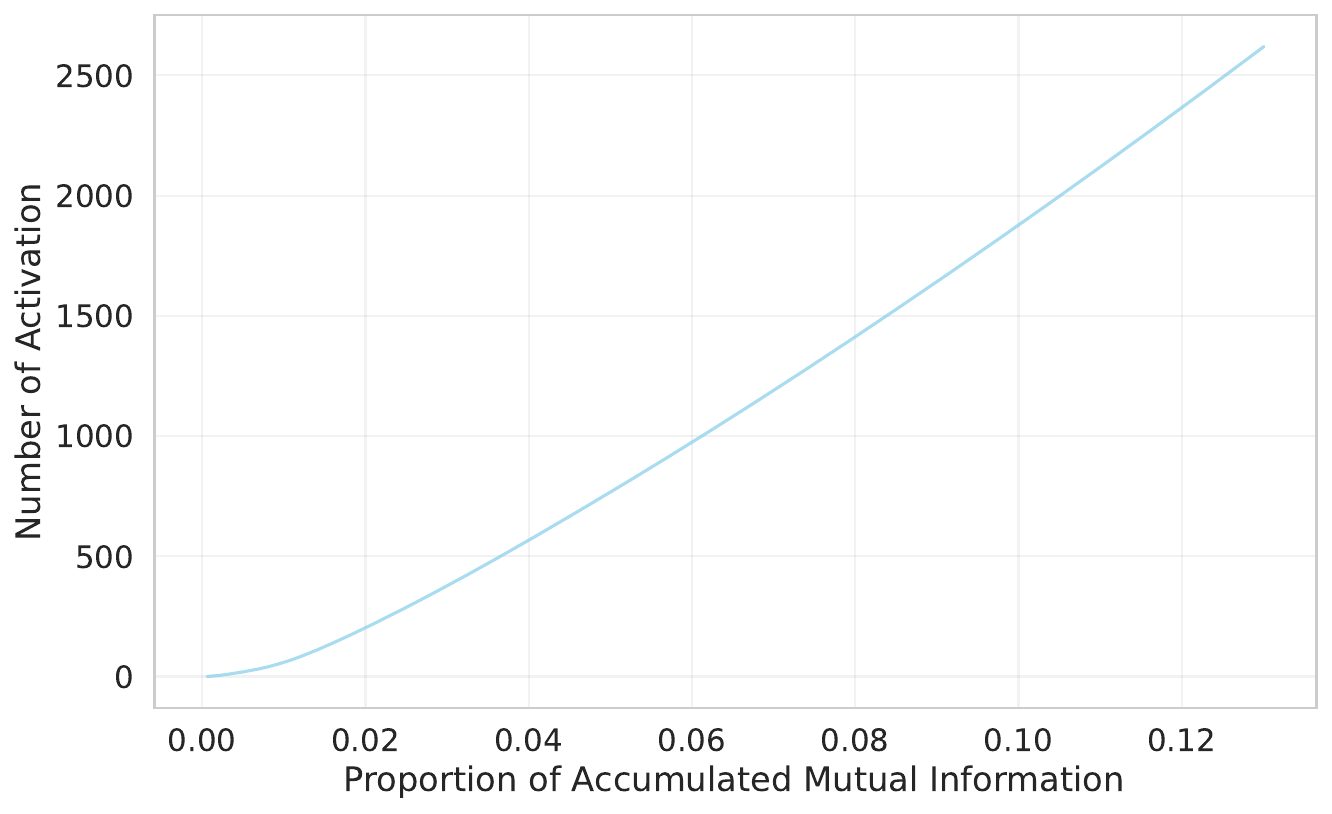}
        \caption{Layer 24}
    \end{subfigure}
    \begin{subfigure}[b]{0.32\textwidth}
        \centering
        \includegraphics[width=\linewidth]{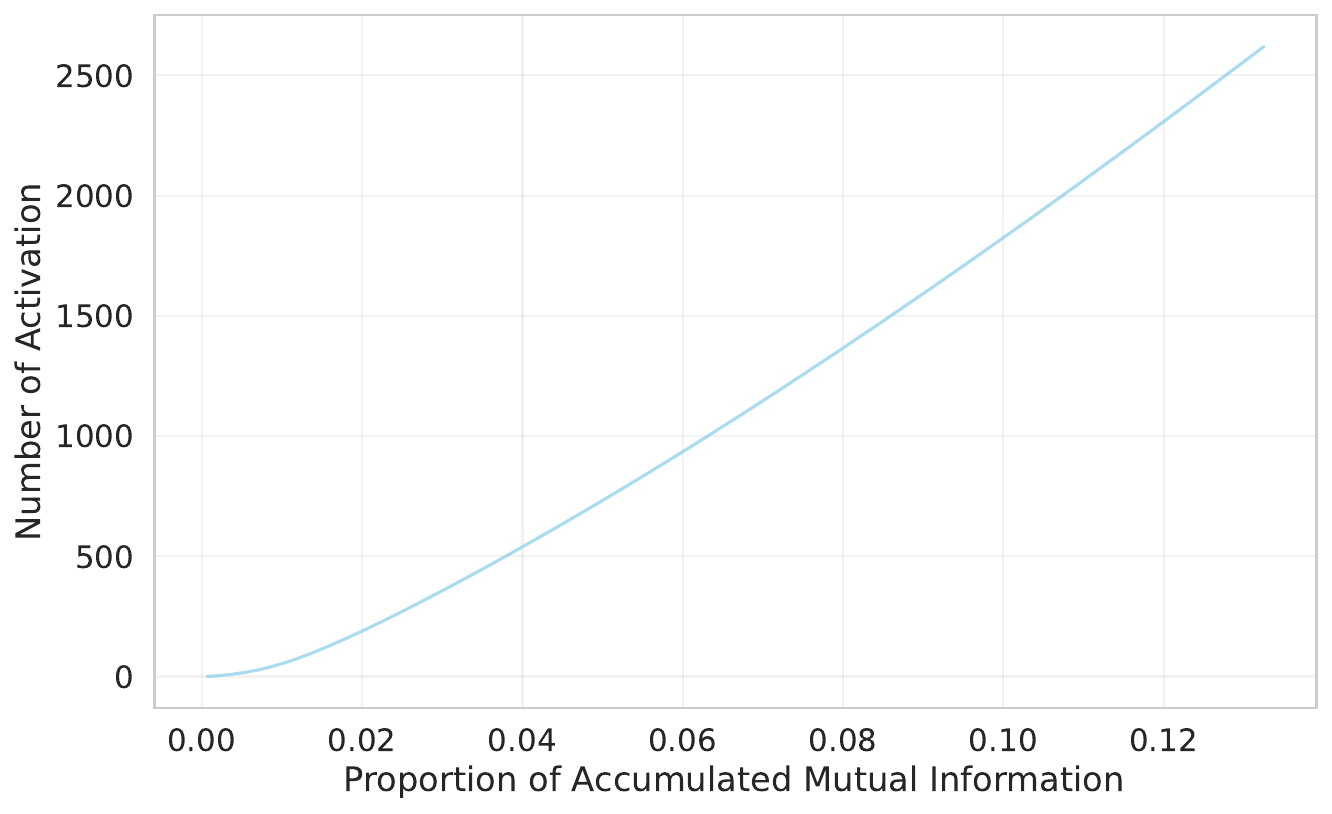}
        \caption{Layer 25}
    \end{subfigure}
\caption{Proportion of accumulated mutual Information ($K$) on Gemma2-9B}
\label{fig:dist_mutual_K_gemma}
\end{figure*}
\begin{figure*}[t]
    \centering
    \begin{subfigure}[b]{0.32\textwidth}
        \centering
        \includegraphics[width=\linewidth]{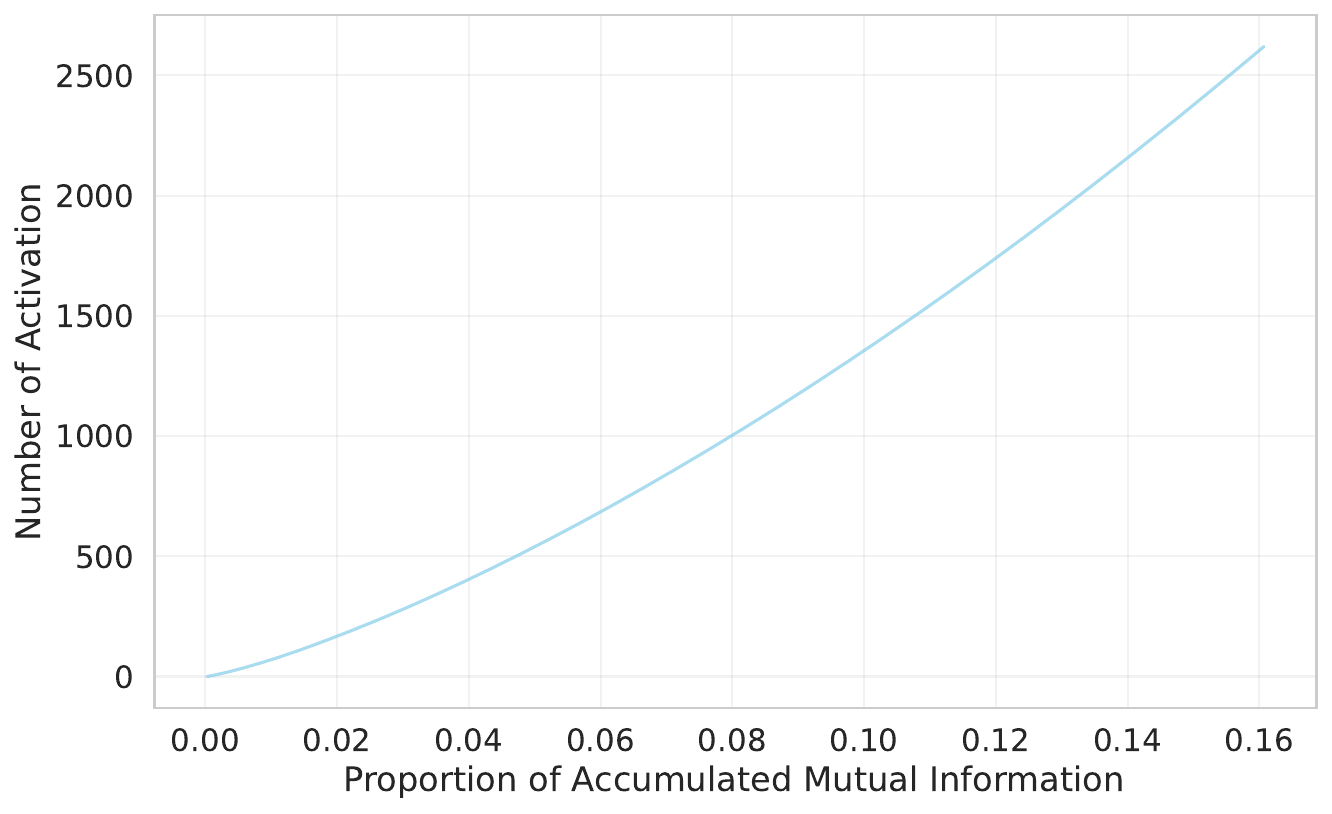}
        \caption{Layer 13}
    \end{subfigure}
    \begin{subfigure}[b]{0.32\textwidth}
        \centering
        \includegraphics[width=\linewidth]{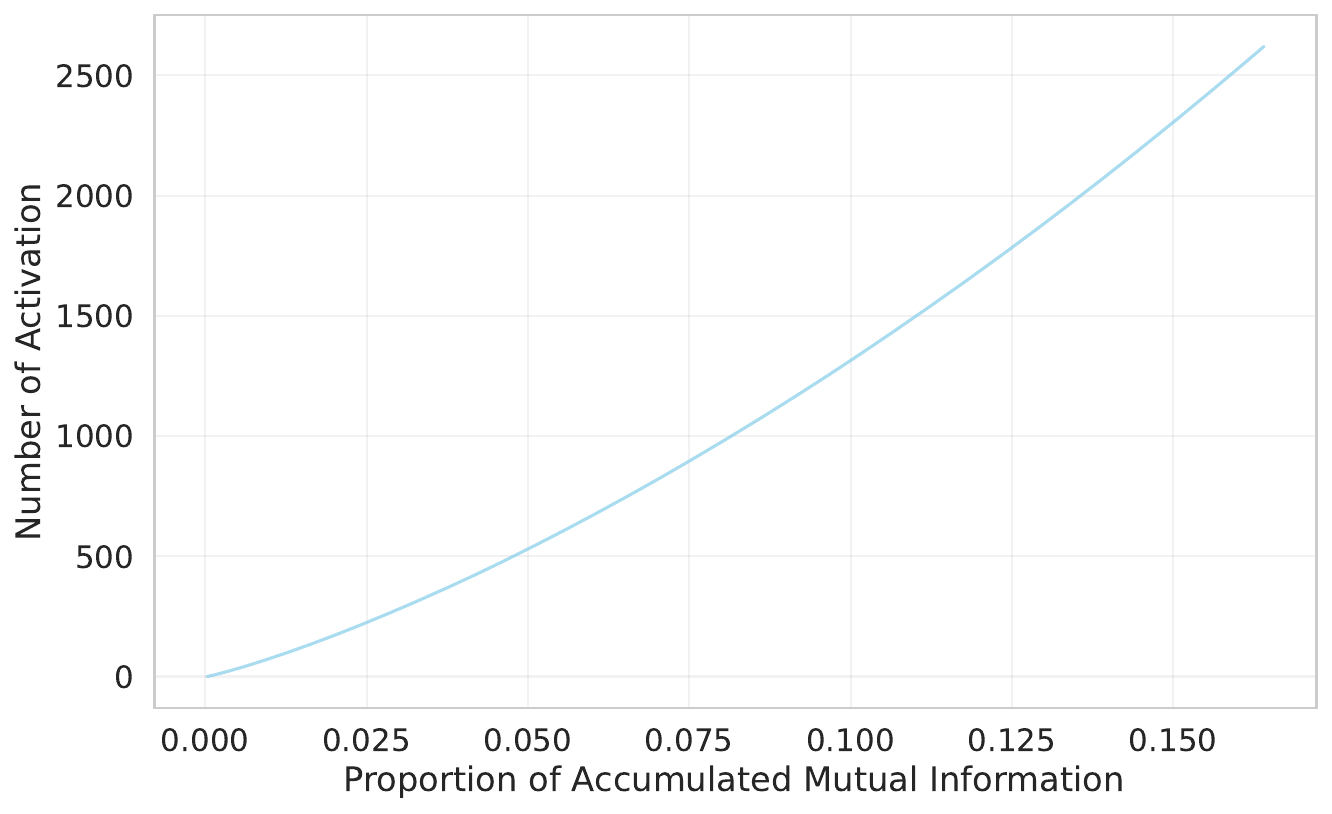}
        \caption{Layer 14}
    \end{subfigure}
    \begin{subfigure}[b]{0.32\textwidth}
        \centering
        \includegraphics[width=\linewidth]{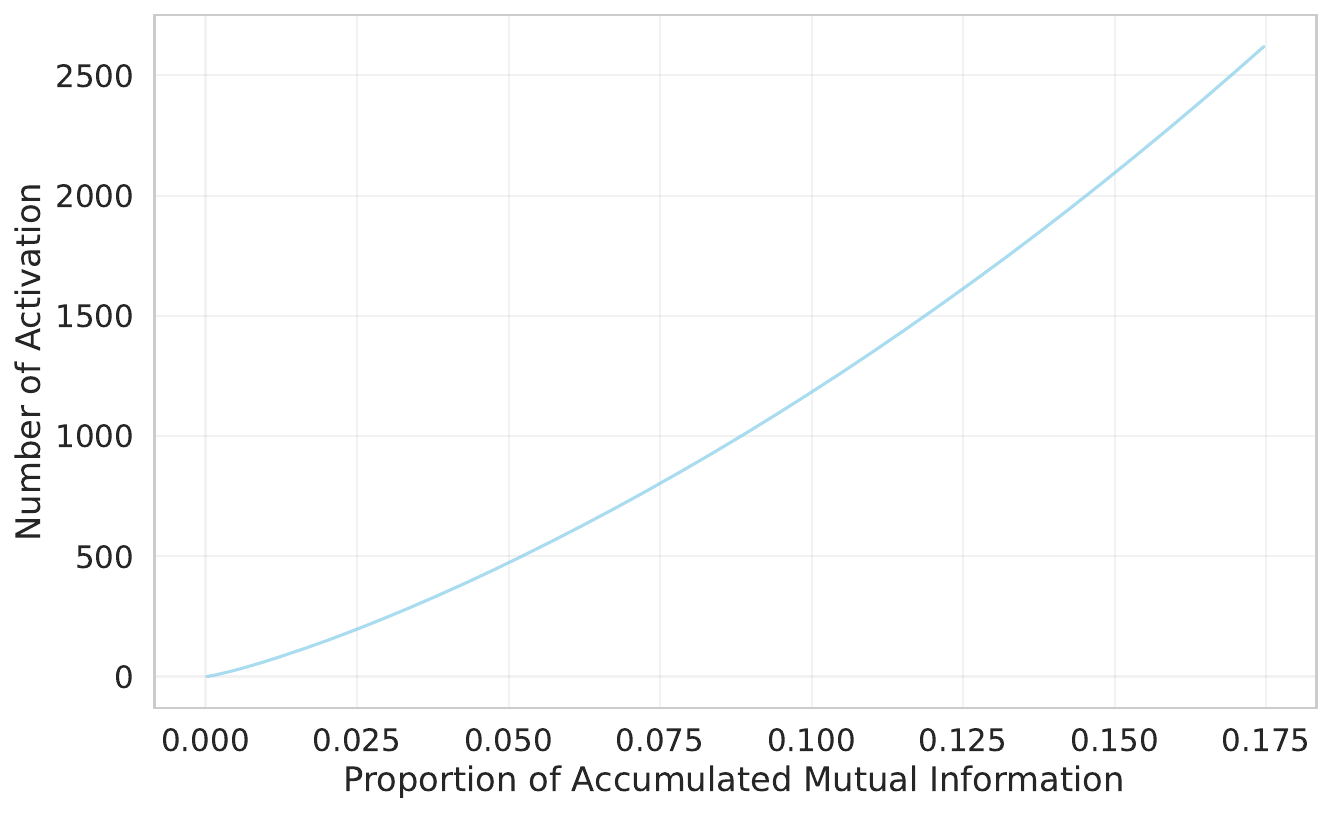}
        \caption{Layer 15}
    \end{subfigure}
\caption{Proportion of accumulated mutual Information ($K$) on Llama2-7B}
\label{fig:dist_mutual_K_ll2}
\end{figure*}
\begin{figure*}[t]
    \centering
    \begin{subfigure}[b]{0.32\textwidth}
        \centering
        \includegraphics[width=\linewidth]{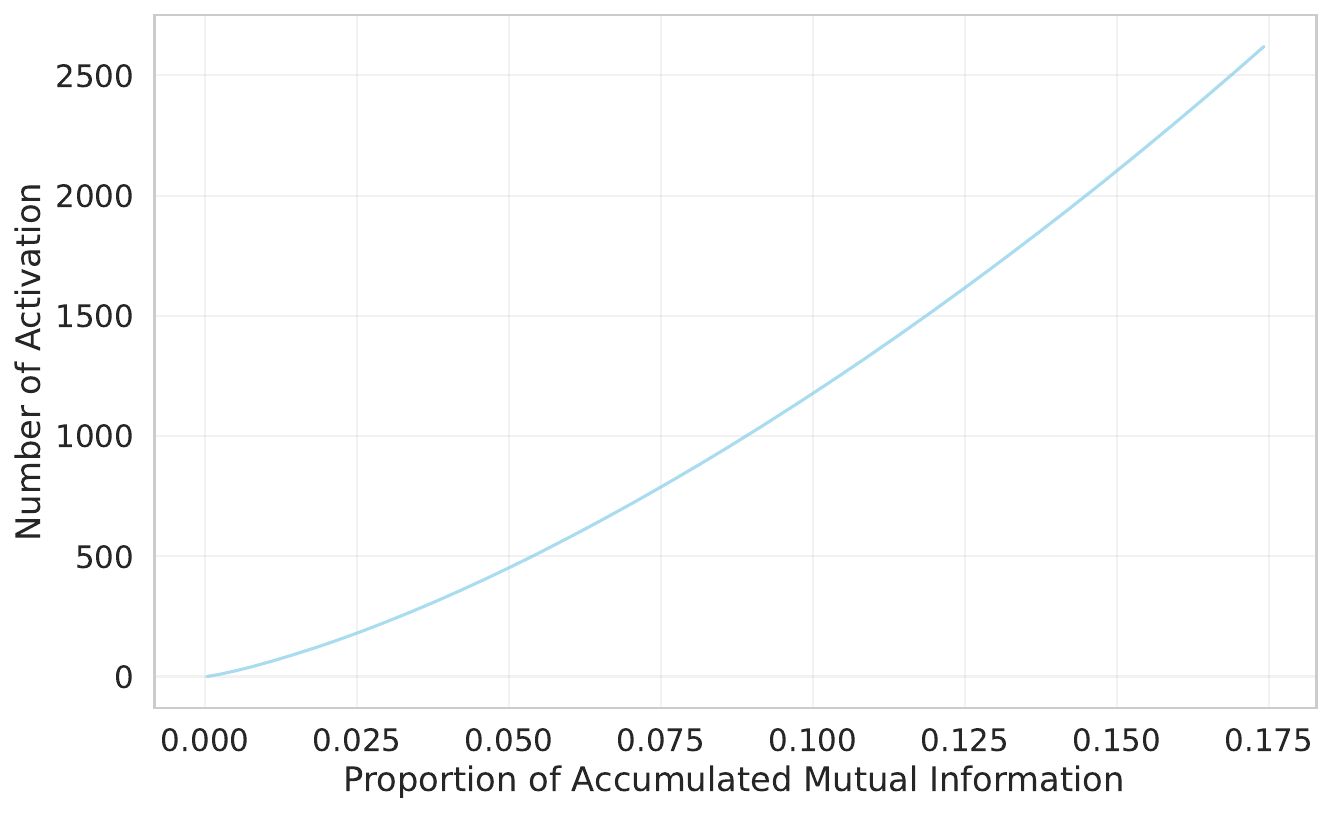}
        \caption{Layer 13}
    \end{subfigure}
    \begin{subfigure}[b]{0.32\textwidth}
        \centering
        \includegraphics[width=\linewidth]{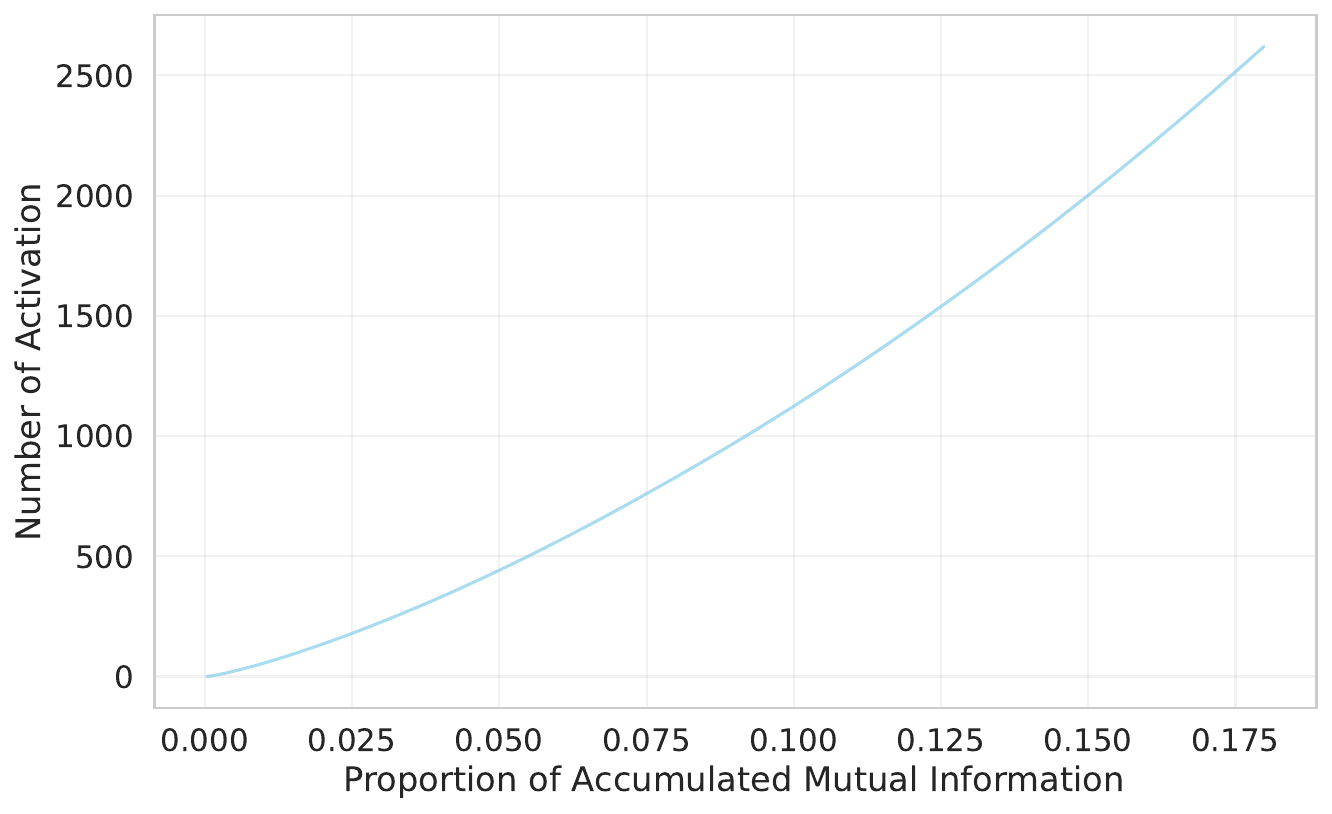}
        \caption{Layer 14}
    \end{subfigure}
    \begin{subfigure}[b]{0.32\textwidth}
        \centering
        \includegraphics[width=\linewidth]{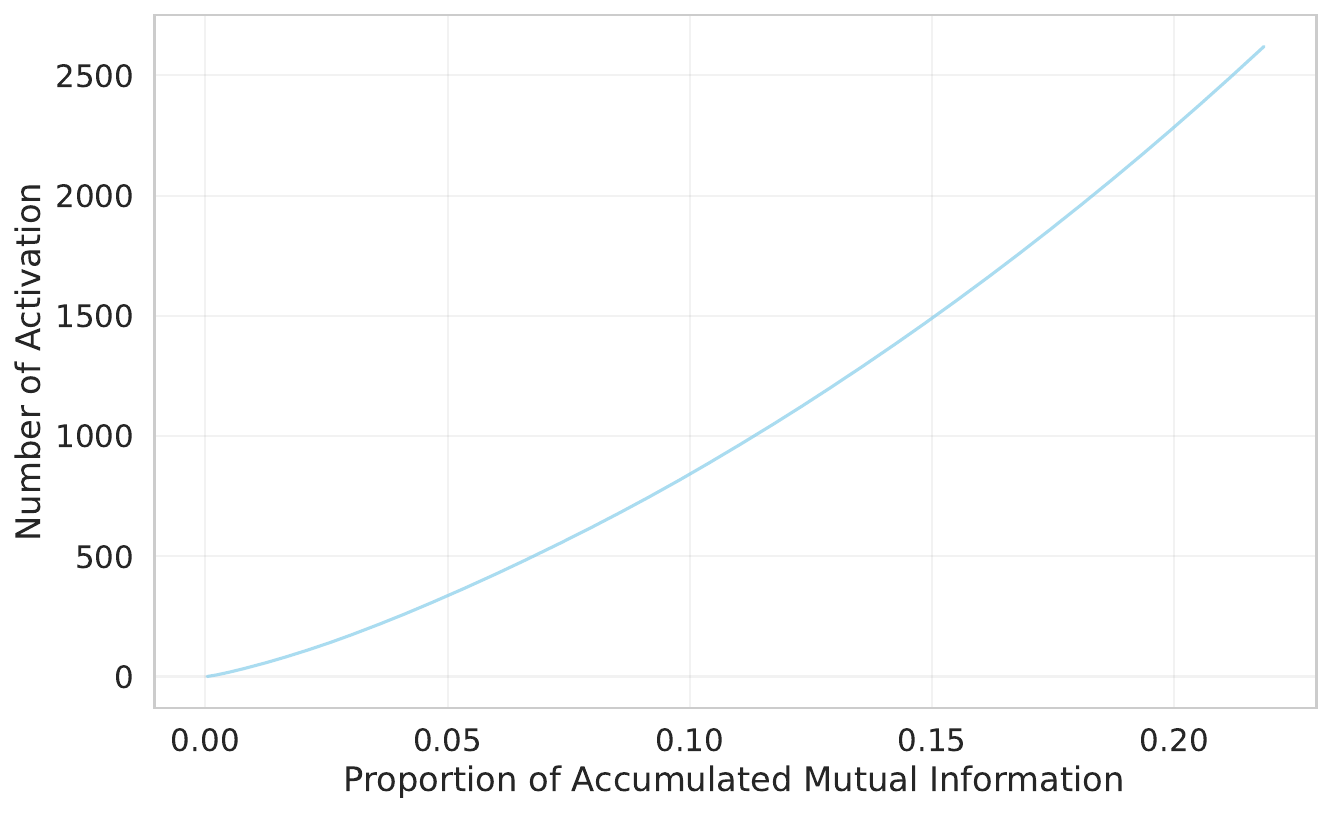}
        \caption{Layer 15}
    \end{subfigure}
\caption{Proportion of accumulated mutual Information ($K$) on Llama3-8B}
\label{fig:dist_mutual_K_ll3}
\end{figure*}
\section{Distribution of Mutual Information}
\label{sec:mutual-information-distribution}
In \cref{sec:detail-indentify-functional-activations}, we mentioned that we use the proportion mutual information ($K$) to determine how many activations of the SAE ($k$) to select. \cref{fig:dist_mutual_K_gemma,fig:dist_mutual_K_ll2,fig:dist_mutual_K_ll3}  shows the layer-wise accumulated mutual information (x-axis) for the number of selected activations (y-axis) across different models (Gemma2-9B, Llama2-7B, Llama3-8B). In all three models, we observed that the graph is skewed when $k$ (y-axis) is small, indicating that some SAE activations have relatively high mutual information. While there were some differences in this tendency between models (particularly pronounced in Gemma2-9B), we found only a little variation across selected layers within the same model. This analysis corresponds to the $K$ values (from 0.01 for Gemma2-9B to 0.07 for Llama3-8B) that we identified through hyperparameter search in \cref{sec:baseline-details}.

\section{Distribution Patterns of the Residual Stream Under Knowledge Conflict}
\label{sec:distribution-pattern}

In this section, we provide further analysis of the representation patterns when knowledge conflicts in addition to~\cref{fig:skewness-change}.
Here, we focus on analysing the distribution patterns of different knowledge selection behaviours.
More specifically, we compare the representation difference between the activation from \UseContextDataset and \UseParameterDataset, which are both under knowledge conflict but select contextual and parametric knowledge to generate the answer \ConflictAnswer and \MemorisedAnswer.
In~\cref{sec:skewness-pattern}, we analyse the skewness patterns; in~\cref{sec:norm-pattern}, we analyse the L1 norm and L2 norm patterns since previous work~\citep{DBLP:journals/corr/abs-2406-11430} also show the norm value may be related to the contextual information usage.
We provide more residual stream analysis in our preliminary study~\citep{residual-stream-knowledge-conflict}.

\subsection{Skewness of Residual Stream}
\label{sec:skewness-pattern}
\begin{figure*}[t]
    \centering
    \begin{subfigure}[b]{0.32\textwidth}
        \centering
        \includegraphics[width=\linewidth]{MINT-figures/Patterns/Llama-2-7b-hf_nqswap_hidden_kurtosis.pdf}
    \end{subfigure}
    \begin{subfigure}[b]{0.32\textwidth}
        \centering
        \includegraphics[width=\linewidth]{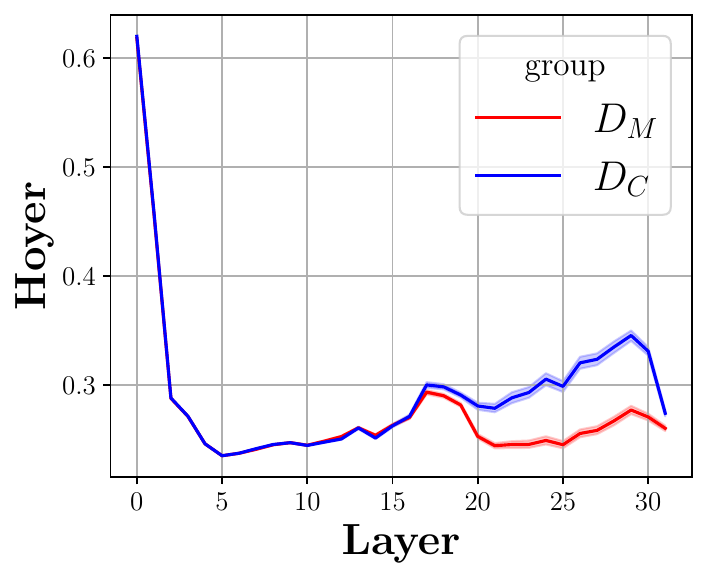}
    \end{subfigure}
    \begin{subfigure}[b]{0.32\textwidth}
        \centering
        \includegraphics[width=\linewidth]{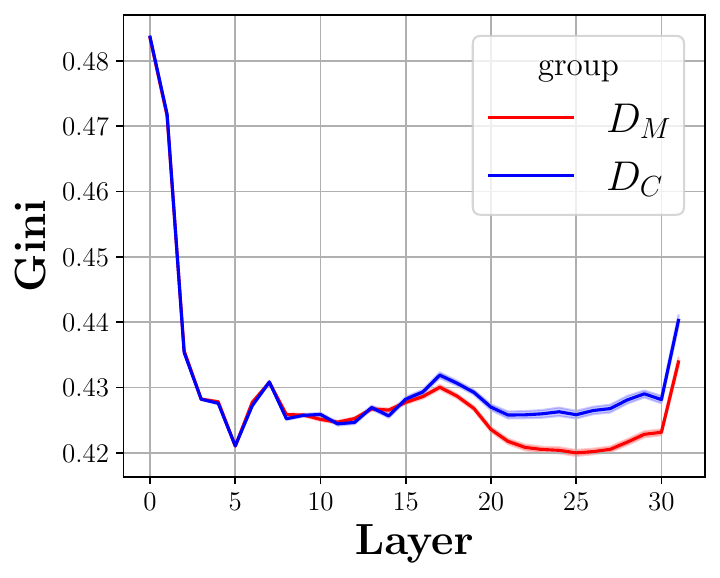}
    \end{subfigure}
\caption{Skewness of the hidden states of Llama2-7B on NQSwap.}
\label{fig:llama2-hidden-skewness}
\end{figure*}

\begin{figure*}[t]
    \centering
    \begin{subfigure}[b]{0.32\textwidth}
        \centering
        \includegraphics[width=\linewidth]{MINT-figures/Patterns/Meta-Llama-3-8B_nqswap_hidden_kurtosis.pdf}
    \end{subfigure}
    \begin{subfigure}[b]{0.32\textwidth}
        \centering
        \includegraphics[width=\linewidth]{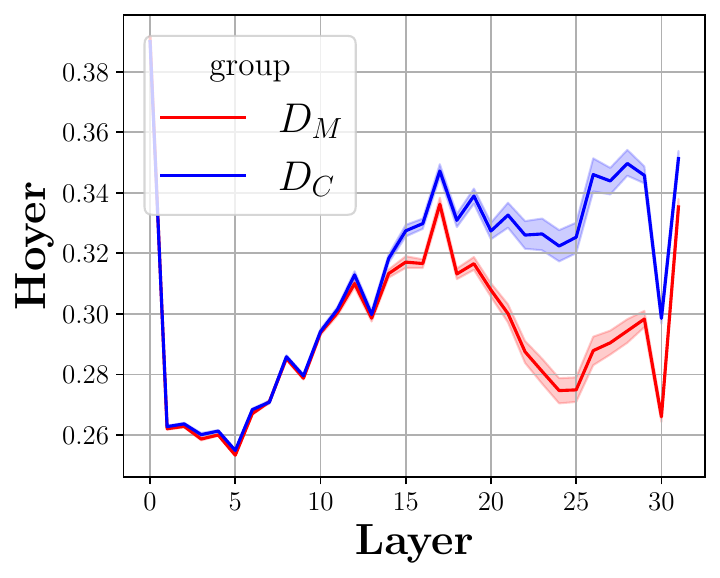}
    \end{subfigure}
    \begin{subfigure}[b]{0.32\textwidth}
        \centering
        \includegraphics[width=\linewidth]{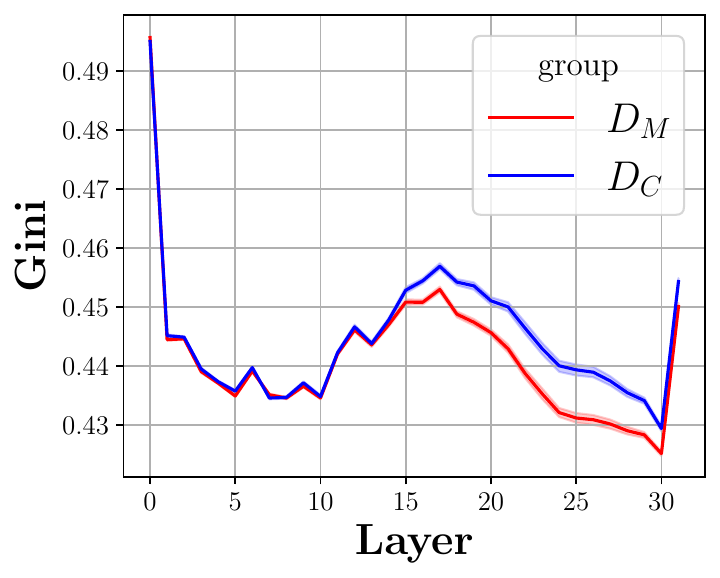}
    \end{subfigure}
\caption{Skewness of the hidden states of Llama3-8B on NQSwap.}
\label{fig:llama3-hidden-skewness}
\end{figure*}

\begin{figure*}[t]
    \centering
    \begin{subfigure}[b]{0.32\textwidth}
        \centering
        \includegraphics[width=\linewidth]{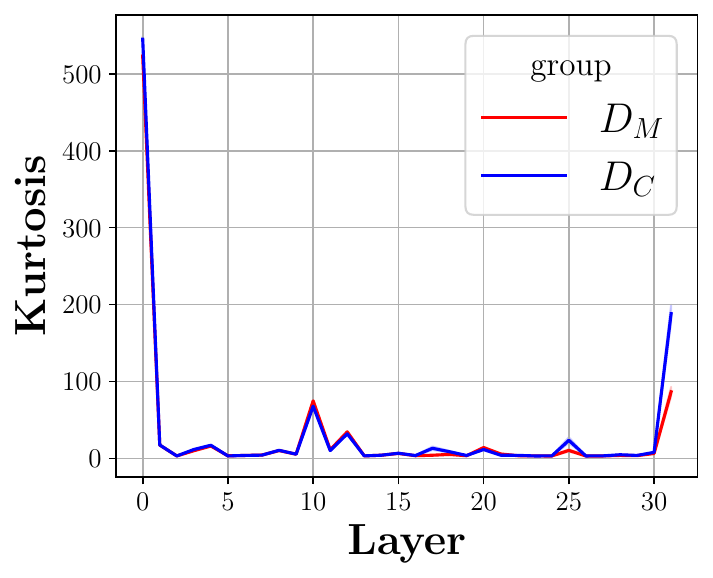}
    \end{subfigure}
    \begin{subfigure}[b]{0.32\textwidth}
        \centering
        \includegraphics[width=\linewidth]{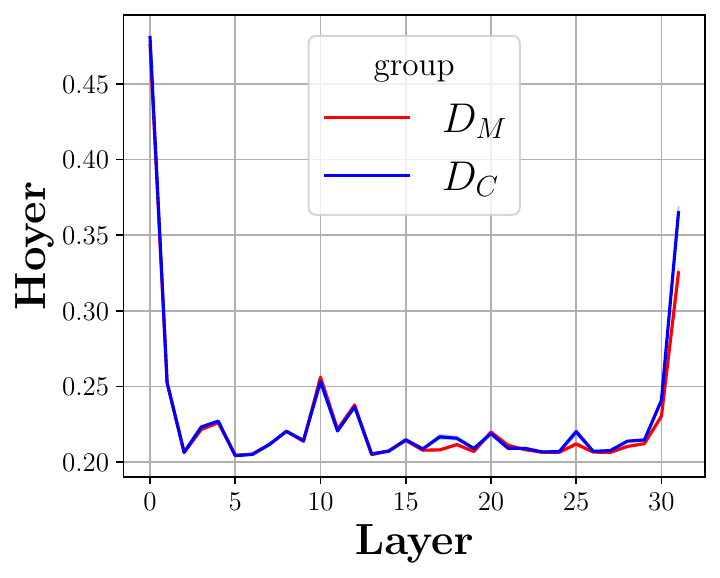}
    \end{subfigure}
    \begin{subfigure}[b]{0.32\textwidth}
        \centering
        \includegraphics[width=\linewidth]{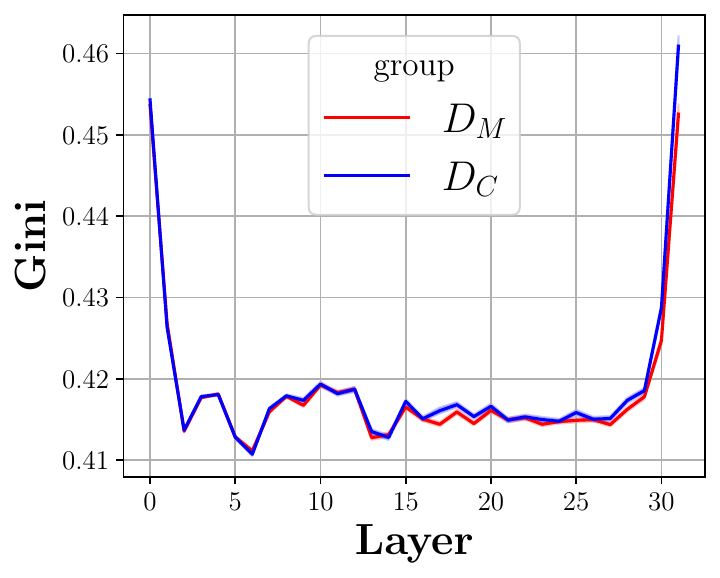}
    \end{subfigure}
\caption{Skewness of the MLP activation of Llama2-7B on NQSwap.}
\label{fig:llama2-mlp-skewness}
\end{figure*}

\begin{figure*}[t]
    \centering
    \begin{subfigure}[b]{0.32\textwidth}
        \centering
        \includegraphics[width=\linewidth]{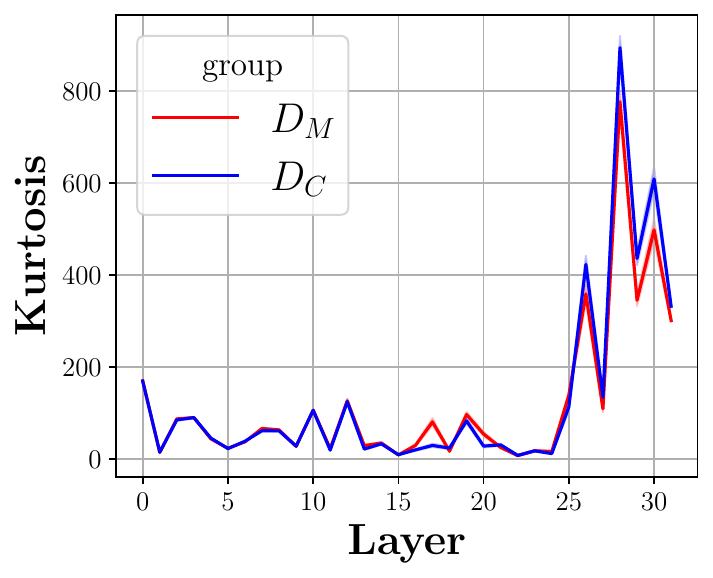}
    \end{subfigure}
    \begin{subfigure}[b]{0.32\textwidth}
        \centering
        \includegraphics[width=\linewidth]{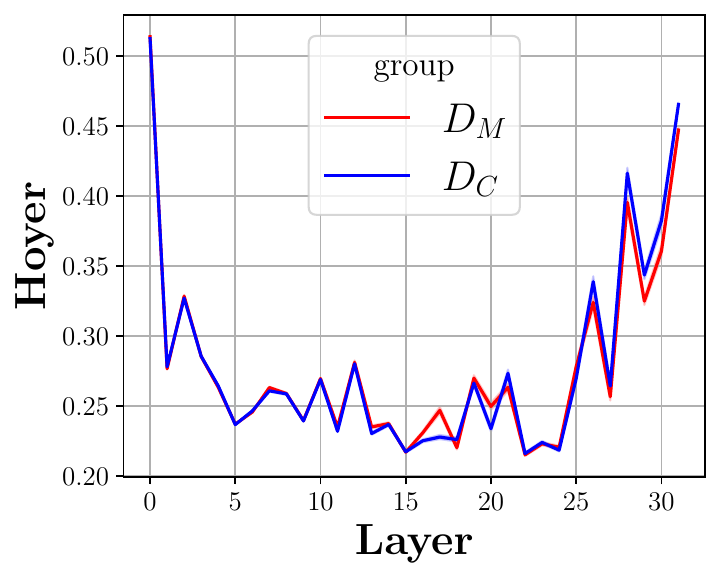}
    \end{subfigure}
    \begin{subfigure}[b]{0.32\textwidth}
        \centering
        \includegraphics[width=\linewidth]{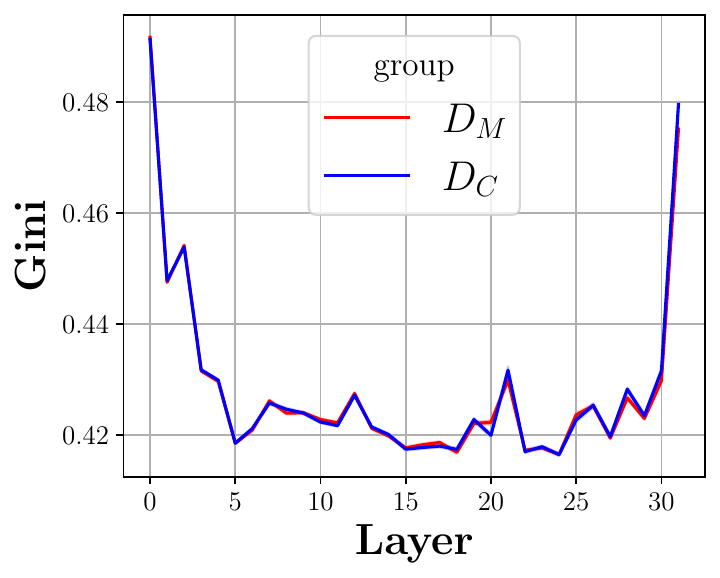}
    \end{subfigure}
\caption{Skewness of the Self-Attention activation of Llama3-8B on NQSwap.}
\label{fig:llama3-attn-skewness}
\end{figure*}

In addition to Kurtosis, we used in~\cref{fig:skewness-change}, we also measure the skewness by Hoyer and Gini index.
We present the skewness patterns of hidden states in~\cref{fig:llama2-hidden-skewness} and~\cref{fig:llama3-hidden-skewness}.
We find the residual stream exhibits a significant skewed pattern when selecting the contextual knowledge to generate the answer.
This observation supports the effectiveness of \MethodName, where the residual steam becomes skewed when \MethodName steers the model to generate contextual knowledge as shown in~\cref{fig:skewness-change}.
We also analyse the skewness pattern of MLP activations and Self-Attention activations in~\cref{fig:llama2-mlp-skewness} and~\cref{fig:llama3-attn-skewness}.
However, we do not observe a distinct distribution difference like hidden states. 

\begin{figure*}[t]
    \centering
    \begin{subfigure}[b]{0.32\textwidth}
        \centering
        \includegraphics[width=\linewidth]{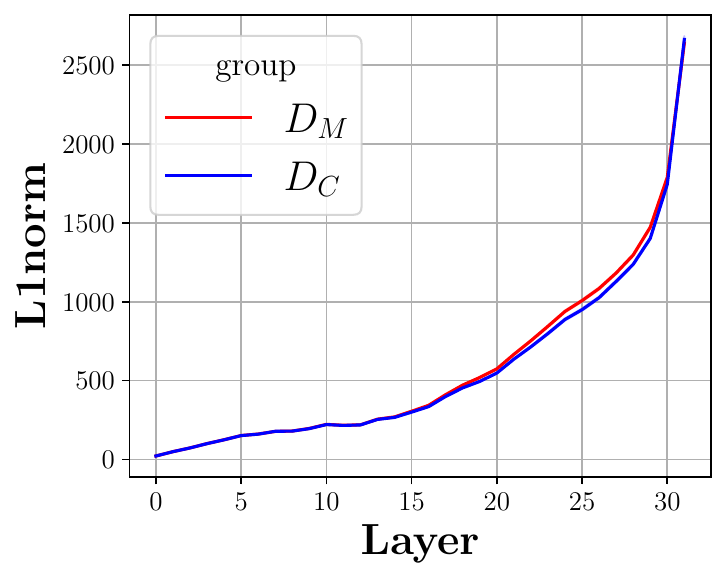}
    \end{subfigure} \qquad
    \begin{subfigure}[b]{0.32\textwidth}
        \centering
        \includegraphics[width=\linewidth]{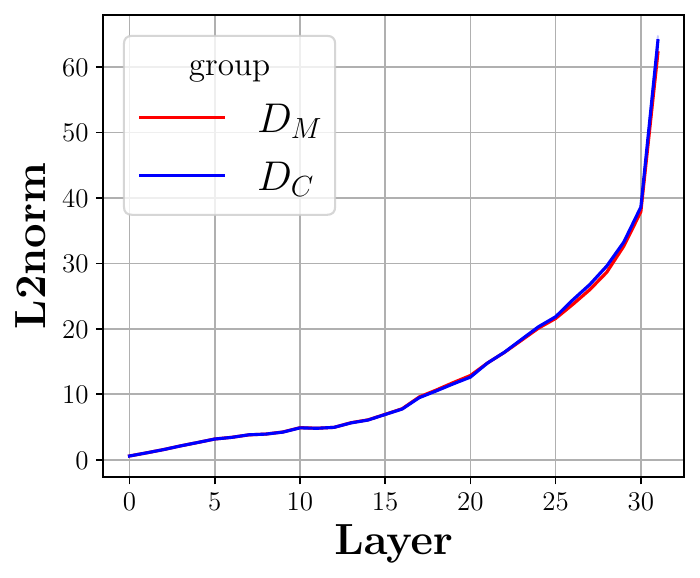}
    \end{subfigure}
\caption{L1 norm and L2 norm of the hidden states of Llama3-8B on NQSwap.}
\label{fig:llama3-norm}
\end{figure*}
\begin{figure*}[t]
    \centering
    \begin{subfigure}[b]{0.32\textwidth}
        \centering
        \includegraphics[width=\linewidth]{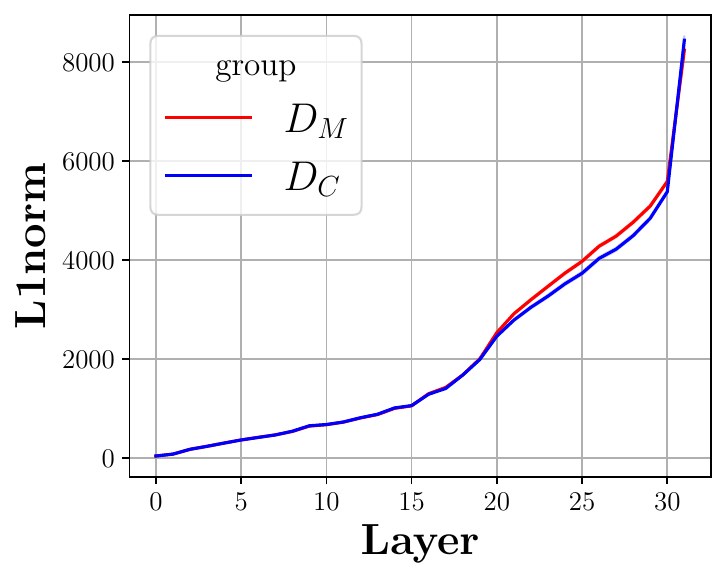}
    \end{subfigure} \qquad
    \begin{subfigure}[b]{0.32\textwidth}
        \centering
        \includegraphics[width=\linewidth]{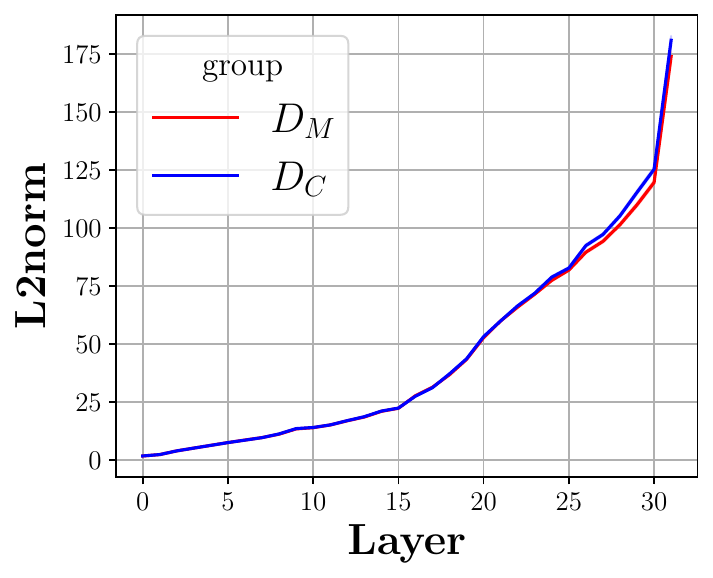}
    \end{subfigure}
\caption{L1 norm and L2 norm of the hidden states of Llama2-7B on NQSwap.}
\label{fig:llama2-norm}
\end{figure*}

\subsection{L1 Norm and L2 Norm Pattern}
\label{sec:norm-pattern}
As we observe the distinct skewness pattern in hidden states, we further analyse their L1 norm and L2 norm patterns in~\cref{fig:llama3-norm} and~\cref{fig:llama2-norm}
However, we do not observe the distinct norm differences between \UseContextDataset and \UseParameterDataset, though they have a significantly different skewness pattern.

\end{document}